\documentclass[1p]{elsarticle}
\usepackage{amsmath,amssymb,subfigure}
\usepackage{floatpag}

\journal{Pattern Recognition}
\floatpagestyle{empty}

\begin{document}

\title{{Moments and Root-Mean-Square Error of the Bayesian MMSE Estimator of
Classification Error in the Gaussian Model}}

\begin{abstract}
The most important aspect of any classifier is its error rate, because this
quantifies its predictive capacity. Thus, the accuracy of error estimation
is critical. Error estimation is problematic in small-sample classifier
design because the error must be estimated using the same data from which
the classifier has been designed. Use of prior knowledge, in the form of a
prior distribution on an uncertainty class of feature-label distributions to
which the true, but unknown, feature-distribution belongs, can facilitate
accurate error estimation (in the mean-square sense) in circumstances where
accurate completely model-free error estimation is impossible. This paper
provides analytic asymptotically exact finite-sample approximations for
various performance metrics of the resulting Bayesian Minimum
Mean-Square-Error (MMSE) error estimator in the case of linear discriminant
analysis (LDA) in the multivariate Gaussian model. These performance metrics
include the first, second, and cross moments of the Bayesian MMSE error
estimator with the true error of LDA, and therefore, the Root-Mean-Square
(RMS) error of the estimator. We lay down the theoretical groundwork for
Kolmogorov double-asymptotics in a Bayesian setting, which enables us to
derive asymptotic expressions of the desired performance metrics. From these
we produce analytic finite-sample approximations and demonstrate their
accuracy via numerical examples. Various examples illustrate the behavior of
these approximations and their use in determining the necessary sample size
to achieve a desired RMS. The Supplementary Material contains derivations
for some equations and added figures.
\end{abstract}

\newtheorem{theorem}{Theorem} \newtheorem{corollary}[theorem]{Corollary} %
\newtheorem{lemma}[theorem]{Lemma}

\begin{frontmatter}

\author[TAMU,STAT]{Amin~Zollanvari\corref{cor1}}
\ead{amin\_zoll@neo.tamu.edu}
\cortext[cor1]{Corresponding author.}
\author[TAMU,TGEN]{Edward R. Dougherty}
\ead{edward@ece.tamu.edu}

\address[TAMU]{Department of Electrical and Computer Engineering,
Texas A\&M University, College Station, TX 77843}
\address[STAT]{Department of Statistics,
Texas A\&M University, College Station, TX 77843}
\address[TGEN]{Translational Genomics Research Institute (TGEN), Phoenix, AZ 85004}


\end{frontmatter}

\section{Introduction}

The most important aspect of any classifier is its error, $\varepsilon $,
defined as the probability of misclassification, since $\varepsilon $
quantifies the predictive capacity of the classifier. Relative to a
classification rule and a given feature-label distribution, the error is a
function of the sampling distribution and as such possesses its own
distribution, which characterizes the true performance of the classification
rule. In practice, the error must be estimated from data by some error
estimation rule yielding an estimate, $\hat{\varepsilon}$. If samples are
large, then part of the data can be held out for error estimation;
otherwise, the classification and error estimation rules are applied on the
same set of training data, which is the situation that concerns us here.
Like the true error, the estimated error is also a function of the sampling
distribution. The performance of the error estimation rule is completely
described by its joint distribution, $(\varepsilon ,\hat{\varepsilon})$.

Three widely-used metrics for performance of an error estimator are the
bias, deviation variance, and root-mean-square (RMS), given by 
\begin{equation}
\begin{aligned}
& \text{Bias}[\hat{\varepsilon}]\,=\,E[\hat{\varepsilon}]-E[\varepsilon
]\,,\;\;\; \\[-0.5ex]
& \text{Var}^{d}[\hat{\varepsilon}]\,=\,\text{Var}(\hat{\varepsilon}%
-\varepsilon )\,=\,\text{Var}(\varepsilon )+\text{Var}(\hat{\varepsilon})-2%
\text{Cov}(\varepsilon ,\hat{\varepsilon})\,, \\[-0.5ex]
& \text{RMS}[\hat{\varepsilon}]\,=\,\sqrt{E[(\varepsilon -\hat{\varepsilon}%
)^{2}]}\,=\,\sqrt{E[\varepsilon ^{2}]+E[{\hat{\varepsilon}}%
^{2}]-2E[\varepsilon \hat{\varepsilon}]}=\sqrt{\text{Bias}[\hat{\varepsilon}%
]^{2}+\text{Var}^{d}[\hat{\varepsilon}]}\,,
\end{aligned}
\label{eq-RMS}
\end{equation}%
respectively. The RMS (square root of mean square error, MSE) is the most
important because it quantifies estimation accuracy. Bias requires only the
first-order moments, whereas the deviation variance and RMS require also the
second-order moments.

Historically, analytic study has mainly focused on the first marginal moment
of the estimated error for linear discriminant analysis (LDA) in the
Gaussian model or for multinomial discrimination ~\cite{Hill:66}-\cite{ZollBragDoug:09}; however,
marginal knowledge does not provide the joint probabilistic knowledge
required for assessing estimation accuracy, in particular, the mixed second
moment. Recent work has aimed at characterizing joint behavior. For
multinomial discrimination, exact representations of the second-order
moments, both marginal and mixed, for the true error and the resubstitution
and leave-one-out estimators have been obtained~\cite{Braga:10}. For LDA,
the exact joint distributions for both resubstitution and leave-one-out have
been found in the univariate Gaussian model and approximations have been
found in the multivariate model with a common known covariance matrix~\cite%
{ZollBragDoug:12,ZollBragDoug:10}. Whereas one could utilize the approximate
representations to find approximate moments via integration in the
multivariate model with a common known covariance matrix, more accurate
approximations, including the second-order mixed moment and the RMS, can be
achieved via asymptotically exact analytic expressions using a double
asymptotic approach, where both sample size ($n$) and dimensionality ($p$)
approach infinity at a fixed rate between the two~\cite{Zollanvari}.
Finite-sample approximations from the double asymptotic method have shown to
be quite accurate \cite{Zollanvari,Wyma:90,pike:76}. There is quite a body
of work on the use of double asymptotics for the analysis of LDA and its
related statistics \cite%
{Zollanvari,Raud:72,Deev:70,Fuji:00,Serd:00,Bickel:04}. Raudys and Young
provide a good review of the literature on the subject \cite{RaudYoun:04}.

Although the theoretical underpinning of both~\cite{Zollanvari} and the
present paper relies on double asymptotic expansions, in which $%
n,p\rightarrow \infty $ at a proportional rate, our practical interest is in
the finite-sample approximations corresponding to the asymptotic expansions.
In \cite{Wyma:90}, the accuracy of such finite-sample approximations was
investigated relative to asymptotic expansions for the expected error of LDA
in a Gaussian model. Several single-asymptotic expansions ($n\rightarrow
\infty $) were considered, along with double-asymptotic expansions ($%
n,p\rightarrow \infty $) \cite{Raud:72,Deev:70}. The results of \cite%
{Wyma:90} show that the double-asymptotic approximations are significantly
more accurate than the single-asymptotic approximations. In particular, even
with $n/p<3$, the double-asymptotic expansions yield \textquotedblleft
excellent approximations\textquotedblright\ while the others
\textquotedblleft falter.\textquotedblright\ 

The aforementioned work is based on the assumption that a sample is drawn
from a fixed feature-label distribution $F$, a classifier and error estimate
are derived from the sample without using any knowledge concerning $F$, and
performance is relative to $F$. Research dating to 1978, shows that
small-sample error estimation under this paradigm tends to be inaccurate.
Re-sampling methods such as cross-validation possess large deviation
variance and, therefore, large RMS \cite{Glic:78,DougCB:09}. Scientific
content in the context of small-sample classification can be facilitated by
prior knowledge \cite{ILLUSION, Lori1, Lori}. There are three possibilities
regarding the feature-label distribution: (1) $F$ is known, in which case no
data are needed and there is no error estimation issue; (2) nothing is known
about $F$, there are no known RMS bounds, or those that are known are
useless for small samples; and (3) $F$ is known to belong to an uncertainty
class of distributions and this knowledge can be used to either bound the
RMS \cite{Zollanvari} or be used in conjunction with the training data to
estimate the error of the designed classifier. If there exists a prior
distribution governing the uncertainty class, then in essence we have a
distributional model. Since virtually nothing can be said about the error
estimate in the first two cases, for a classifier to possess scientific
content we must begin with a distributional model.

Given the need for a distributional model, a natural approach is to find an
optimal minimum mean-square-error (MMSE) error estimator relative to an
uncertainty class $\Theta $~\cite{Lori1}. This results in a Bayesian
approach with $\Theta $ being given a prior distribution, $\pi (\theta
),\theta \in \Theta $, and the sample $S_{n}$ being used to construct a
posterior distribution, $\pi ^{\ast }(\theta )$, from which an optimal MMSE
error estimator, $\hat{\varepsilon}^{B}$, can be derived. $\pi (\theta )$
provides a mathematical framework for both the analysis of any error
estimator and the design of estimators with desirable properties or optimal
performance. $\pi ^{\ast }(\theta )$ provides a sample-conditioned
distribution on the true classifier error, where randomness in the true
error comes from uncertainty in the underlying feature-label distribution
(given $S_{n}$). Finding the sample-conditioned MSE, $\text{MSE}_{\theta }[%
\hat{\varepsilon}^{B}|S_{n}]$, of an MMSE error estimator amounts to
evaluating the variance of the true error conditioned on the observed sample~%
\cite{Lori2}. $\text{MSE}_{\theta }[\hat{\varepsilon}^{B}|S_{n}]\rightarrow 0
$ as $n\rightarrow \infty $ almost surely in both the discrete and Gaussian
models provided in~\cite{Lori2,Lori3}, where closed form expressions for the
sample-conditioned MSE are available.

The sample-conditioned MSE provides a measure of performance across the
uncertainty class $\Theta $ for a given sample $S_{n}$. As such, it involves
various sample-conditioned moments for the error estimator: $E_{\theta }[%
\hat{\varepsilon}^{B}|S_{n}]$, $E_{\theta }[(\hat{\varepsilon}%
^{B})^{2}|S_{n}]$, and $E_{\theta }[\varepsilon \hat{\varepsilon}^{B}|S_{n}]$%
. One could, on the other hand, consider the MSE relative to a fixed
feature-label distribution in the uncertainty class and randomness relative
to the sampling distribution. This would yield the
feature-label-distribution-conditioned MSE, $\text{MSE}_{S_{n}}[\hat{%
\varepsilon}^{B}|\theta ]$, and the corresponding moments: $E_{S_{n}}[\hat{%
\varepsilon}^{B}|\theta ]$, $E_{S_{n}}[(\hat{\varepsilon}^{B})^{2}|\theta ]$%
, and $E_{S_{n}}[\varepsilon \hat{\varepsilon}^{B}|\theta ]$. From a
classical point of view, the moments given $\theta $ are of interest as they
help shed light on the performance of an estimator relative to fixed
parameters of class conditional densities. Using this set of moments (i.e.
given $\theta $) we are able to compare the performance of the Bayesian MMSE
error estimator to classical estimators of true error such as resubstitution
and leave-one-out.

From a global perspective, to evaluate performance across both the
uncertainty class and the sampling distribution requires the unconditioned
MSE, $\text{MSE}_{\theta ,S_{n}}[\hat{\varepsilon}^{B}]$, and corresponding
moments $E_{\theta ,S_{n}}[\hat{\varepsilon}^{B}]$, $E_{\theta ,S_{n}}[(\hat{%
\varepsilon}^{B})^{2}]$, and $E_{\theta ,S_{n}}[\varepsilon \hat{\varepsilon}%
^{B}]$. While both $\text{MSE}_{S_{n}}[\hat{\varepsilon}^{B}|\theta ]$ and $%
\text{MSE}_{\theta ,S_{n}}[\hat{\varepsilon}^{B}]$ have been examined via
simulation studies in \cite{Lori1,Lori,Lori3} for discrete and Gaussian
models, our intention in the present paper is to derive double-asymptotic
representations of the feature-labeled conditioned (given $\theta $) and
unconditioned MSE, along with the corresponding moments of the Bayesian MMSE
error estimator for linear discriminant analysis (LDA) in the Gaussian model.

We make three modeling assumptions. As in many analytic error analysis
studies, we employ stratified sampling: $n=n_{0}+n_{1}$ sample points are
selected to constitute the sample $S_{n}$ in $R^{p}$, where given $n$, $n_{0}
$ and $n_{1}$ are determined, and where $\mathbf{x}_{1},\mathbf{x}%
_{2},\ldots ,\mathbf{x}_{n_{0}}$ and $\mathbf{x}_{n_{0}+1},\mathbf{x}%
_{n_{0}+2},...$ $,\mathbf{x}_{n_{0}+n_{1}}$\ are randomly selected from
distributions $\Pi _{0}$ and $\Pi _{1}$, respectively. $\Pi _{i}$ possesses
a multivariate Gaussian distribution $N(\mathbf{\boldsymbol{\mu }}_{i},%
\mathbf{\Sigma })$, for $i=0,1$. This means that the prior probabilities $%
\alpha _{0}$ and $\alpha _{1}=1-\alpha _{0}$ for classes 0 and 1,
respectively, cannot be estimated from the sample (see \cite{Ande:51} for a
discussion of issues surrounding lack of an estimator for $\alpha _{0}$).
However, our second assumption is that $\alpha _{0}$ and $\alpha _{1}$ are
known. This is a natural assumption for many medical classification
problems. If we desire early or mid-term detection, then we are typically
constrained to a small sample for which $n_{0}$ and $n_{1}$ are not random
but for which $\alpha _{0}$ and $\alpha _{1}$ are known (estimated with
extreme accuracy) on account of a large population of post-mortem
examinations. The third assumption is that there is a known common
covariance matrix for the classes, a common assumption in error analysis 
\cite{John:61,Soru:71,Mora:75,Zollanvari}. The common covariance assumption
is typical for small samples because it is well-known that LDA commonly
performs better that quadratic discriminant analysis (QDA) for small
samples, even if the actual covariances are different, owing to the
estimation advantage of using the pooled sample covariance matrix. As for
the assumption of known covariance, this assumption is typical simply owing
to the mathematical difficulties of obtaining error representations for
unknown covariance (we know of no unknown-covariance result for second-order
representations). Indeed, the natural next step following this paper and 
\cite{Zollanvari} is to address the unknown covariance problem (although
with it being outstanding for almost half a century, it may prove difficult).

Under our assumptions, the \emph{Anderson} $W$ \emph{statistic} is defined
by 
\begin{equation}
W(\bar{\mathbf{\mathbf{x}}}_{0},\bar{\mathbf{x}}_{1},\mathbf{x})\,=\,\left( 
\mathbf{x}-\frac{\bar{\mathbf{x}}_{0}+\bar{\mathbf{x}}_{1}}{2}\right) ^{T}%
\mathbf{\Sigma }^{-1}\left( \bar{\mathbf{x}}_{0}-\bar{\mathbf{x}}_{1}\right)
,  \label{LDAc}
\end{equation}%
where $\bar{\mathbf{x}}_{0}=\frac{1}{n_{0}}\sum_{i=1}^{n_{0}}\mathbf{x}_{i}$
and $\bar{\mathbf{x}}_{1}=\frac{1}{n_{1}}\sum_{i=n_{0}+1}^{n_{0}+n_{1}}%
\mathbf{x}_{i}$. The corresponding linear discriminant is defined by $\psi
_{n}(\mathbf{x})=1$ if $W(\bar{\mathbf{x}}_{0},\bar{\mathbf{x}}_{1},\mathbf{x%
})\leq c$ and $\psi _{n}(\mathbf{x})=0$ if \vspace{-0.2cm} $W(\bar{\mathbf{x}%
}_{0},\bar{\mathbf{x}}_{1},\mathbf{x})>c$, where $c=\log \frac{1-\alpha _{0}%
}{\alpha _{0}}$. Given sample $S_{n}$ (and thus $\bar{\mathbf{x}}_{0}$ and $%
\bar{\mathbf{x}}_{1}$), for $i=0,1$, the error for $\psi _{n}$ is given by $%
\varepsilon =\alpha _{0}\varepsilon _{0}+\alpha _{1}\varepsilon _{1}$, where 
\begin{equation}
\begin{aligned}
\varepsilon _{i}=\Phi \left( \frac{(-1)^{i+1}\left( \boldsymbol{\mu }_{i}-%
\frac{\bar{\mathbf{x}}_{0}+\bar{\mathbf{x}}_{1}}{2}\right) ^{T}\mathbf{%
\Sigma }^{-1}\left( \bar{\mathbf{x}}_{0}-\bar{\mathbf{x}}_{1}\right)
+(-1)^{i}c}{\sqrt{{(\bar{\mathbf{x}}_{0}-\bar{\mathbf{x}}_{1})}^{T}\mathbf{%
\Sigma }^{-1}{(\bar{\mathbf{x}}_{0}-\bar{\mathbf{x}}_{1})}}}\right) 
\end{aligned}
\label{eq:true_errorU}
\end{equation}%
and $\Phi (.)$ denotes the cumulative distribution function of a standard
normal random variable.

Raudys proposed the following approximation to the expected LDA
classification error \cite{Raud:72,RaudYoun:04}: 
\begin{equation}
\begin{aligned}
E_{S_{n}}[{\varepsilon }_{0}]=P(W(\bar{\mathbf{x}}_{0},\bar{\mathbf{x}}_{1},%
\mathbf{x})\leq c\mid \mathbf{x}\in \Pi _{0})\,\eqsim \,\Phi \left( \frac{%
-E_{S_{n}}[W(\bar{\mathbf{x}}_{0},\bar{\mathbf{x}}_{1},\mathbf{x})\mid 
\mathbf{x}\in \Pi _{0}]+c}{\sqrt{\text{Var}_{S_{n}}[W(\bar{\mathbf{x}}_{0},%
\bar{\mathbf{x}}_{1},\mathbf{x})\mid \mathbf{x}\in \Pi _{0}]}}\right) \,
\end{aligned}%
\end{equation}%
\bigskip We provide similar approximations for error-estimation moments and
prove asymptotic exactness.

\section{Bayesian MMSE Error Estimator}

In the Bayesian classification framework \cite{Lori1, Lori}, it is assumed
that the class-0 an class-1 conditional distributions are parameterized by $%
\theta _{0}$ and $\theta _{1}$, respectively. Therefore, assuming known $%
\alpha_i$, the actual feature-label distribution belongs to an uncertainty
class parameterized by $\theta =(\theta _{0},\theta _{1})$ according to a
prior distribution, $\pi (\theta )$. Given a sample $S_{n}$, the Bayesian
MMSE error estimator minimizes the MSE between the true error of a designed
classifier, $\psi _{n}$, and an error estimate (a function of $S_{n}$ and $%
\psi _{n}$). The expectation in the MSE is taken over the uncertainty class
conditioned on $S_{n}$. Specifically, the MMSE error estimator is the
expected true error, $\hat{\varepsilon}^{B}(S_{n})=\mathrm{E}_{\theta
}[\varepsilon (\theta )|S_{n}]$. The expectation given the sample is over
the posterior density, $\pi ^{\ast }(\theta )$. Thus, we write the Bayesian
MMSE error estimator as $\hat{\varepsilon}^{B}=\mathrm{E}_{\pi ^{\ast
}}[\varepsilon ]$. To facilitate analytic representations, $\theta _{0}$ and 
$\theta _{1}$ are assumed to be independent prior to observing the data.
Denote the marginal priors of $\theta _{0} $ and $\theta _{1}$ by $\pi
(\theta _{0})$ and $\pi (\theta _{1})$, respectively, and the corresponding
posteriors by $\pi ^{\ast }(\theta _{0})$ and $\pi ^{\ast }(\theta _{1})$,
respectively. Independence is preserved, i.e., $\pi ^{\ast }(\theta
_{0},\theta _{1})=\pi ^{\ast }(\theta _{0})\pi ^{\ast }(\theta _{1})$ for $%
i=0,1$~\cite{Lori1}.

Owing to the posterior independence between $\theta _{0}$ and $\theta _{1}$
and the fact that $\alpha _{i}$ is known, the Bayesian MMSE error estimator
can be expressed by~\cite{Lori1} 
\begin{equation}
\hat{\varepsilon}^{B}=\alpha _{0}\mathrm{E}_{\pi ^{\ast }}[\varepsilon
_{0}]+\alpha _{1}\mathrm{E}_{\pi ^{\ast }}[\varepsilon _{1}]=\alpha _{0}\hat{%
\varepsilon}_{0}^{B}+\alpha _{1}\hat{\varepsilon}_{1}^{B},  \label{eq:BEEU}
\end{equation}%
where, letting $\mathbf{\Theta }_{i}$ be the parameter space of $\theta _{i}$%
, 
\begin{equation}
\hat{\varepsilon}_{i}^{B}=\mathrm{E}_{\pi ^{\ast }}[\varepsilon _{i}]=\int_{%
\mathbf{\Theta }_{i}}\varepsilon _{i}(\theta _{i})\pi ^{\ast }(\theta
_{i})d\theta _{i}.  \label{eq:MMSE:general}
\end{equation}%
For known $\mathbf{\Sigma }$ and the prior distribution on $\boldsymbol{\mu }%
_{i}$ assumed to be Gaussian with mean $\mathbf{m}_{i}$ and covariance
matrix $\mathbf{\Sigma }/\nu _{i}$, $\hat{\varepsilon}_{i}^{B}$ is given by
equation (10) in \cite{Lori}: 
\begin{equation}
\begin{aligned}
\hat{\varepsilon}_{i}^{B}=\Phi \left( (-1)^{i}\;\frac{-\left( \mathbf{m}%
_{i}^{\ast }-\frac{\bar{\mathbf{x}}_{0}+\bar{\mathbf{x}}_{1}}{2}\right) ^{T}%
\mathbf{\Sigma }^{-1}\left( \bar{\mathbf{x}}_{0}-\bar{\mathbf{x}}_{1}\right)
+c}{\sqrt{{(\bar{\mathbf{x}}_{0}-\bar{\mathbf{x}}_{1})}^{T}\mathbf{\Sigma }%
^{-1}{(\bar{\mathbf{x}}_{0}-\bar{\mathbf{x}}_{1})}}}\sqrt{\frac{\nu
_{i}^{\ast }}{\nu _{i}^{\ast }+1}}\right) ,
\end{aligned}
\label{qwsaqwsaU}
\end{equation}%
where 
\begin{equation}
\begin{aligned}
\mathbf{m}_{i}^{\ast }=\frac{n_{i}\bar{\mathbf{x}}_{i}+\nu _{i}\mathbf{m}_{i}%
}{n_{i}+\nu _{i}},\quad \nu _{i}^{\ast }=n_{i}+\nu _{i}\,.
\end{aligned}
\label{mnuU}
\end{equation}%
and $\nu _{i}>0$ is a measure {of our certainty concerning the prior knowedge%
} -- the larger $\nu _{i}$ is the more localized the prior distribution is
about $\mathbf{m}_{i}$. Letting $\boldsymbol{\mu }=[\boldsymbol{\mu }%
_{0}^{T},\boldsymbol{\mu }_{1}^{T}]^{T}$, the moments that interest us are
of the form $E_{S_{n}}[\hat{\varepsilon}^{B}|\boldsymbol{\mu }]$, $%
E_{S_{n}}[(\hat{\varepsilon}^{B})^{2}|\boldsymbol{\mu }]$, and $%
E_{S_{n}}[\varepsilon \hat{\varepsilon}^{B}|\boldsymbol{\mu }]$, which are
used to obtain $\text{MSE}_{S_{n}}[\hat{\varepsilon}^{B}|\boldsymbol{\mu }]$%
, and $E_{\boldsymbol{\mu },S_{n}}[\hat{\varepsilon}^{B}]$, $E_{\boldsymbol{%
\mu },S_{n}}[(\hat{\varepsilon}^{B})^{2}]$, and $E_{\boldsymbol{\mu }%
,S_{n}}[\varepsilon \hat{\varepsilon}^{B}]$, which are needed to
characterize $\text{MSE}_{\boldsymbol{\mu },S_{n}}[\hat{\varepsilon}^{B}]$.

\section{Bayesian-Kolmogorov Asymptotic Conditions}

The Raudys-Kolmogorov asymptotic conditions \cite{Zollanvari} are defined on
a sequence of Gaussian discrimination problems with a sequence of parameters
and sample sizes: $(\boldsymbol{\mu }_{p,0},\boldsymbol{\mu }_{p,1},\mathbf{%
\Sigma }_{p},n_{p,0},n_{p,1})$, $p=1,2,\ldots $, where the means and the
covariance matrix are arbitrary. The common assumptions for
Raudys-Kolmogorov asymptotics are $n_{0}\rightarrow \infty ,n_{1}\rightarrow
\infty ,p\rightarrow \infty ,\frac{p}{n_{0}}\rightarrow J_{0}<\infty ,\frac{p%
}{n_{1}}\rightarrow J_{1}<\infty $. For notational simplicity, we denote the
limit under these conditions by $\underset{{p\rightarrow \infty }}{\operatorname{lim}%
}$. In the analysis of classical statistics related to LDA it is commonly
assumed that the Mahalanobis distance, $\delta _{\boldsymbol{\mu },{p}}\,=\,%
\sqrt{(\boldsymbol{\mu }_{p,0}-\boldsymbol{\mu }_{p,1})^{T}\mathbf{\Sigma }%
_{p}^{-1}(\boldsymbol{\mu }_{p,0}-\boldsymbol{\mu }_{p,1})}$, is finite and $%
\underset{{p\rightarrow \infty }}{\operatorname{lim}}\delta _{\boldsymbol{\mu },p}=%
\overline{\delta }_{\boldsymbol{\mu }}$ (see \cite{Serd:00}, p. 4). This
condition assures existence of limits of performance metrics of the relevant
statistics \cite{Zollanvari,Serd:00}.

To analyze the Bayesian MMSE error estimator, $\hat{\varepsilon}_{i}^{B}$,
we modify the sequence of Gaussian discrimination problems to: 
\begin{equation}
(\boldsymbol{\mu }_{p,0},\boldsymbol{\mu }_{p,1},\mathbf{\Sigma }%
_{p},n_{p,0},n_{p,1},\mathbf{m}_{p,0},\mathbf{m}_{p,1},\nu _{p,0},\nu
_{p,1})\,,p=1,2,\ldots   \label{gkusguk}
\end{equation}%
In addition to the previous conditions, we assume that the following limits
exist for $i,j=0,1$: $\underset{{p\rightarrow \infty }}{\operatorname{lim}}\mathbf{m}%
_{p,i}^{T}\mathbf{\Sigma }_{p}^{-1}\boldsymbol{\mu }_{p,j}=\overline{\mathbf{%
m}_{i}^{T}\mathbf{\Sigma }^{-1}\boldsymbol{\mu }_{j}}$, $\underset{{%
p\rightarrow \infty }}{\operatorname{lim}}\mathbf{m}_{p,i}^{T}\mathbf{\Sigma }%
_{p}^{-1}\mathbf{m}_{p,j}=\overline{\mathbf{m}_{i}^{T}\mathbf{\Sigma }^{-1}%
\mathbf{m}_{j}}$, and $\underset{{p\rightarrow \infty }}{\operatorname{lim}}%
\boldsymbol{\mu }_{p,i}^{T}\mathbf{\Sigma }_{p}^{-1}\boldsymbol{\mu }_{p,j}=%
\overline{\boldsymbol{\mu }_{i}^{T}\mathbf{\Sigma }^{-1}\boldsymbol{\mu }_{j}%
}$, where $\overline{\mathbf{m}_{i}^{T}\mathbf{\Sigma }^{-1}\boldsymbol{\mu }%
_{j}}$, $\overline{\mathbf{m}_{i}^{T}\mathbf{\Sigma }^{-1}\mathbf{m}_{j}}$,
and $\overline{\boldsymbol{\mu }_{i}^{T}\mathbf{\Sigma }^{-1}\boldsymbol{\mu 
}_{j}}$ are some constants to which the limits converge. In \cite{Serd:00},
fairly mild sufficient conditions are given for the existence of these
limits.

We refer to all of the aforementioned conditions, along with $\nu
_{i}\rightarrow \infty ,\frac{\nu _{i}}{n_{i}}\rightarrow \gamma _{i}<\infty 
$, as the \emph{Bayesian-Kolmogorov asymptotic conditions} (b.k.a.c). We
denote the limit under these conditions by $\lim_{\substack{ \text{b.k.a.c.} 
\\ }}$, which means that, for $i,j=0,1$, 
\begin{equation}
\lim_{\substack{ \text{b.k.a.c.}  \\ }}(\cdot )=\lim_{\substack{ {%
p\rightarrow \infty ,n_{i}\rightarrow \infty ,\nu _{i}\rightarrow \infty } 
\\ {\frac{p}{n_{0}}\rightarrow J_{0},\frac{p}{n_{1}}\rightarrow J_{1}},{%
\frac{\nu _{0}}{n_{0}}\rightarrow \gamma _{0},\frac{\nu _{1}}{n_{1}}%
\rightarrow \gamma _{1}}  \\ \gamma _{i}<\infty ,\;J_{i}<\infty  \\ {\mathbf{%
m}_{p,i}^{T}\mathbf{\Sigma }_{p}^{-1}\boldsymbol{\mu }_{p,j}=O(1),\;\;%
\mathbf{m}_{p,i}^{T}\mathbf{\Sigma }_{p}^{-1}\boldsymbol{\mu }%
_{p,j}\rightarrow \overline{\mathbf{m}_{i}^{T}\mathbf{\Sigma }^{-1}%
\boldsymbol{\mu }_{j}}}  \\ \mathbf{m}_{p,i}^{T}\mathbf{\Sigma }_{p}^{-1}%
\mathbf{m}_{p,j}=O(1),\;\;\mathbf{m}_{p,i}^{T}\mathbf{\Sigma }_{p}^{-1}%
\mathbf{m}_{p,j}\rightarrow \overline{\mathbf{m}_{i}^{T}\mathbf{\Sigma }^{-1}%
\mathbf{m}_{j}}  \\ {\boldsymbol{\mu }_{p,i}^{T}\mathbf{\Sigma }_{p}^{-1}%
\boldsymbol{\mu }_{p,j}=O(1),\;\;\boldsymbol{\mu }_{p,i}^{T}\mathbf{\Sigma }%
_{p}^{-1}\boldsymbol{\mu }_{p,j}\rightarrow \overline{\boldsymbol{\mu }%
_{i}^{T}\mathbf{\Sigma }^{-1}\boldsymbol{\mu }_{j}}}}}(\cdot )  \label{KACLU}
\end{equation}%
This limit is defined for the case where there is conditioning on a specific
value of $\boldsymbol{\mu }_{p,i}$. Therefore, in this case $\boldsymbol{\mu 
}_{p,i}$ is not a random variable, and for each $p$, it is a vector of
constants. Absent such conditioning, the sequence of discrimination problems
and the above limit reduce to 
\begin{equation}
(\mathbf{\Sigma }_{p},n_{p,0},n_{p,1},\mathbf{m}_{p,0},\mathbf{m}_{p,1},\nu
_{p,0},\nu _{p,1})\,,p=1,2,\ldots  \label{gkusgukll}
\end{equation}%
and 
\begin{equation}
\lim_{\substack{ \text{b.k.a.c.}  \\ }}(\cdot )=\lim_{\substack{ {%
p\rightarrow \infty ,n_{i}\rightarrow \infty ,\nu _{i}\rightarrow \infty } 
\\ {\frac{p}{n_{0}}\rightarrow J_{0},\frac{p}{n_{1}}\rightarrow J_{1}},{%
\frac{\nu _{0}}{n_{0}}\rightarrow \gamma _{0},\frac{\nu _{1}}{n_{1}}%
\rightarrow \gamma _{1}}  \\ \gamma _{i}<\infty ,\;J_{i}<\infty  \\ \mathbf{m%
}_{p,i}^{T}\mathbf{\Sigma }_{p}^{-1}\mathbf{m}_{p,j}=O(1),\;\;\mathbf{m}%
_{p,i}^{T}\mathbf{\Sigma }_{p}^{-1}\mathbf{m}_{p,j}\rightarrow \overline{%
\mathbf{m}_{i}^{T}\mathbf{\Sigma }^{-1}\mathbf{m}_{j}}}}(\cdot )
\label{KACL2U}
\end{equation}%
respectively. For notational simplicity we assume clarity from the context
and do not explicitly differentiate between these conditions. We denote
convergence in probability under Bayesian-Kolmogorov asymptotic conditions
by \textquotedblleft $\underset{\text{b.k.a.c.}}{\operatorname{plim}}$".
\textquotedblleft $\underset{\text{b.k.a.c.}}{\operatorname{lim}}$" and
\textquotedblleft ${\overset{K}{\rightarrow }}$" denote ordinary convergence
under Bayesian-Kolmogorov asymptotic conditions. At no risk of ambiguity, we
henceforth omit the subscript \textquotedblleft $p$\textquotedblright\ from
the parameters and sample sizes in (\ref{gkusguk}) or (\ref{gkusgukll}).

We define $\eta _{\mathbf{a}_{1},\mathbf{a}_{2},\mathbf{a}_{3},\mathbf{a}%
_{4}}=(\mathbf{a}_{1}-\mathbf{a}_{2})^{T}\mathbf{\Sigma }^{-1}(\mathbf{a}%
_{3}-\mathbf{a}_{4})$ and, for ease of notation write $\eta _{\mathbf{a}_{1},%
\mathbf{a}_{2},\mathbf{a}_{1},\mathbf{a}_{2}}$ as $\eta _{\mathbf{a}_{1},%
\mathbf{a}_{2}}$. There are two special cases: (1) the square of the
Mahalanobis distance in the space of the parameters of the unknown class
conditional densities, $\delta _{\boldsymbol{\mu }}^{2}\,=\,\eta _{%
\boldsymbol{\mu }_{0},\boldsymbol{\mu }_{1}}>0$; and (2) a measure of distance of prior distributions, $\Delta _{\mathbf{m}%
}^{2}\,=\,\eta _{\mathbf{m}_{0},\mathbf{m}_{1}}>0$, where $\mathbf{m}=[%
\mathbf{m}_{0}^{T},\mathbf{m}_{1}^{T}]^{T}$. The conditions in (\ref{KACLU})
assure existence of $\lim_{\mathrm{b.k.a.c}}\eta _{\mathbf{a}_{1},\mathbf{a}%
_{2},\mathbf{a}_{3},\mathbf{a}_{4}}$, where the $\mathbf{a}_{j}$'s can be
any combination of $\mathbf{m}_{i}$ and $\boldsymbol{\mu }_{i}$, $i=0,1$.
Consistent with our notations, we use $\overline{\eta }_{\mathbf{a}_{1},%
\mathbf{a}_{2},\mathbf{a}_{3},\mathbf{a}_{4}}$, $\overline{\delta }_{%
\boldsymbol{\mu }}^{2}$, and $\overline{\Delta }_{\mathbf{m}}^{2}$ to denote
the $\lim_{\mathrm{b.k.a.c}}$ of $\eta _{\mathbf{a}_{1},\mathbf{a}_{2},%
\mathbf{a}_{3},\mathbf{a}_{4}}$, $\delta _{\boldsymbol{\mu }}^{2}$, and $%
\Delta _{\mathbf{m}}^{2}$, respectively. Thus, 
\begin{equation}
\begin{aligned}
\overline{\eta }_{\mathbf{a}_{1},\mathbf{a}_{2},\mathbf{a}_{3},\mathbf{a}%
_{4}}=\overline{(\mathbf{a}_{1}-\mathbf{a}_{2})^{T}\mathbf{\Sigma }^{-1}(%
\mathbf{a}_{3}-\mathbf{a}_{4})}=\overline{\mathbf{a}_{1}^{T}\mathbf{\Sigma }%
^{-1}\mathbf{a}_{3}}-\overline{\mathbf{a}_{1}^{T}\mathbf{\Sigma }^{-1}%
\mathbf{a}_{4}}-\overline{\mathbf{a}_{2}^{T}\mathbf{\Sigma }^{-1}\mathbf{a}%
_{3}}+\overline{\mathbf{a}_{2}^{T}\mathbf{\Sigma }^{-1}\mathbf{a}_{4}}.
\end{aligned}%
\end{equation}

The ratio $p/n_{i}$ is an indicator of complexity for LDA (in fact, any
linear classification rule): the VC dimension in this case is $p+1$ \cite%
{DevrGyorLugo:96}. Therefore, the conditions (\ref{KACLU}) assure the
existence of the asymptotic complexity of the problem. The ratio $\nu
_{i}/n_{i}$ is an indicator of relative certainty of prior knowledge to the
data: the smaller $\nu _{i}/n_{i}$, the more we rely on the data and less on
our prior knowledge. Therefore, the conditions (\ref{KACLU}) state
asymptotic existence of relative certainty. In the following, we let $\beta
_{i}=\frac{\nu_{i}}{n_{i}}$, so that $\beta _{i}=\frac{\nu _{i}}{n_{i}}%
\rightarrow \gamma _{i}$.

\section{First Moment of $\hat{\protect\varepsilon} _{i}^{B}$}

In this section we use the Bayesian-Kolmogorov asymptotic conditions to
characterize the conditional and unconditional first moment of the Bayesian
MMSE error estimator.

\subsection{Conditional Expectation of\textbf{\ $\hat{\protect\varepsilon}%
_{i}^{B}$:\ $E_{S_{n}}[\hat{\protect\varepsilon}_{i}^{B}|\boldsymbol{\protect%
\mu }]$}}

The asymptotic (in a Bayesian-Kolmogorov sense) conditional expectation of
the Bayesian MMSE error estimator is characterized in the following theorem,
with the proof presented in the Appendix. Note that $G_{0}^{B}$, $G_{1}^{B}$%
, and $D$ depend on $\boldsymbol{\mu }$, but to ease the notation we leave
this implicit.

\begin{theorem}
\label{thm-m1} Consider the sequence of Gaussian discrimination problems
defined by (\ref{gkusguk}). Then 
\begin{equation}
\begin{aligned}
\lim_{\substack{ \text{b.k.a.c.}  \\ {}}}E_{S_{n}}[\hat{\varepsilon}
_{i}^{B}|\boldsymbol{\mu}]=\Phi \left( (-1)^{i}\;\frac{-G_{i}^B+c}{\sqrt{D}}%
\right)\,,
\end{aligned}
\label{eq-t1}
\end{equation}%
so that \vspace{-0.2cm} 
\begin{equation}
\lim_{\substack{ \text{b.k.a.c.}  \\ {}}}E_{S_{n}}[\hat{\varepsilon}^{B}|%
\boldsymbol{\mu}]\,={\alpha}_0\Phi \left( \frac{-G_{0}^B+c}{\sqrt{D}}\right)
+{\alpha}_1\Phi \left( \frac{G_{1}^B-c}{\sqrt{D}}\right)\,,  \label{csxasas}
\end{equation}%
where \vspace{-0.2cm} 
\begin{equation}
\begin{aligned}
& G_{0}^B= \frac{1}{2(1+\gamma_0)}\Big( \gamma_0(\overline{\eta}_{ \mathbf{m}%
_0,\boldsymbol{\mu}_1}-\overline{\eta}_{ \mathbf{m}_0,\boldsymbol{\mu}_0})+%
\overline{\delta}^2_{\boldsymbol{\mu}}+(1-\gamma_0)J_0+(1+\gamma_0)J_1 \Big),
\\[-0.6ex]
& G_{1}^B=\frac{-1}{2(1+\gamma_1)}\Big( \gamma_1(\overline{\eta}_{ \mathbf{m}%
_1,\boldsymbol{\mu}_0}-\overline{\eta}_{ \mathbf{m}_1,\boldsymbol{\mu}_1})+%
\overline{\delta}^2_{\boldsymbol{\mu}}+(1-\gamma_1)J_1+(1+\gamma_1)J_0 \Big)
\\[-0.4ex]
& D\,=\,\overline{\delta}^{2}_{\boldsymbol{\mu}}+J_{0}+J _{1}\,. \quad \blacksquare
\end{aligned}
\label{njzz}
\end{equation}
\end{theorem}

Theorem~\ref{thm-m1} suggests a finite-sample approximation: 
\begin{equation}
\begin{aligned}
E_{S_{n}}[\hat{\varepsilon}_{0}^{B}|\boldsymbol{\mu }]\,\eqsim \Phi \left( 
\frac{-G_{0}^{B,f}+c}{\sqrt{\delta _{\boldsymbol{\mu }}^{2}+\frac{p}{n_{0}}+%
\frac{p}{n_{1}}}}\right) ,
\end{aligned}
\label{eq-tdEe0s2dU}
\end{equation}%
where $G_{0}^{B,f}$ is obtained by using the finite-sample parameters of the
problem in (\ref{njzz}), namely, 
\begin{equation}
\begin{aligned}
G_{0}^{B,f}=\frac{1}{2(1+\beta _{0})}\bigg(\beta _{0}(\eta _{\mathbf{m}_{0},%
\boldsymbol{\mu }_{1}}-\eta _{\mathbf{m}_{0},\boldsymbol{\mu }_{0}})+\delta
_{\boldsymbol{\mu }}^{2}+(1-\beta _{0})\frac{p}{n_{0}}+(1+\beta _{0})\frac{p%
}{n_{1}}\bigg).
\end{aligned}
\label{g0f}
\end{equation}%
To obtain the corresponding approximation for $E_{S_{n}}[\hat{\varepsilon}%
_{1}^{B}|\boldsymbol{\mu }]$, it suffices to use (\ref{eq-tdEe0s2dU}) by changing the sign of $c$, 
exchanging $n_{0}$ and $n_{1}$, $\nu _{0}$ and $\nu _{1}$, $\mathbf{m}_{0}$
and $\mathbf{m}_{1}$, and $\boldsymbol{\mu }_{0}$ and $\boldsymbol{\mu }_{1}$
in $-G_{0}^{B,f}$.

To obtain a Raudys-type of finite-sample approximation for the expectation
of $\hat{\varepsilon}_{0}^{B}$, first note that the Gaussian distribution in
(\ref{qwsaqwsaU}) can be rewritten as 
\begin{equation}
\begin{aligned}
\hat{\varepsilon}_{0}^{B}=P(U_{0}(\bar{\mathbf{x}}_{0},\bar{\mathbf{x}}_{1},%
\mathbf{z})\leq c\,|\bar{\mathbf{x}}_{0},\bar{\mathbf{x}}_{1},\mathbf{z}\in
\Psi _{0}),
\end{aligned}
\label{rewriteGaussian}
\end{equation}%
where $\mathbf{z}$ is independent of $S_{n}$, $\Psi _{i}$ is a multivariate
Gaussian $N(\mathbf{m}_{i},\frac{(n_{i}+\nu _{i}+1)(n_{i}+\nu _{i})}{\nu
_{i}^{2}}\mathbf{\Sigma })$, and 
\begin{equation}
\begin{aligned}
U_{i}(\bar{\mathbf{x}}_{0},\bar{\mathbf{x}}_{1},\mathbf{z})=\left( \frac{\nu
_{i}}{n_{i}+\nu _{i}}\mathbf{z}+\frac{n_{i}\bar{\mathbf{x}}_{i}}{n_{i}+\nu
_{i}}-\frac{\bar{\mathbf{x}}_{0}+\bar{\mathbf{x}}_{1}}{2}\right) ^{T}\mathbf{%
\mathbf{\Sigma }}^{-1}\left( \bar{\mathbf{x}}_{0}-\bar{\mathbf{x}}%
_{1}\right) .
\end{aligned}
\label{Ui}
\end{equation}%
Taking the expectation of $\hat{\varepsilon}_{0}^{B}$ relative to the
sampling distribution and then applying the standard normal approximation
yields the Raudys-type of approximation: 
\begin{equation}
\begin{aligned}
E_{S_{n}}[\hat{\varepsilon}_{0}^{B}|\boldsymbol{\mu }]=P(U_{0}(\bar{\mathbf{x%
}}_{0},\bar{\mathbf{x}}_{1},\mathbf{z})\leq c|\mathbf{z}\in \Psi _{0},%
\boldsymbol{\mu })\eqsim \,\Phi \left( \frac{-E_{S_{n},\mathbf{z}}[U_{0}(%
\bar{\mathbf{x}}_{0},\bar{\mathbf{x}}_{1},\mathbf{z})|\mathbf{z}\in \Psi
_{0},\boldsymbol{\mu })]+c}{\sqrt{\text{Var}_{S_{n},\mathbf{z}}[U_{0}(\bar{%
\mathbf{x}}_{0},\bar{\mathbf{x}}_{1},\mathbf{z})|\mathbf{z}\in \Psi _{0},%
\boldsymbol{\mu }]}}\right) .
\end{aligned}
\label{kshbxkabk}
\end{equation}%
Algebraic manipulation yields (Suppl. Section A)

\begin{equation}
\begin{aligned}
E_{S_{n}}[\hat{\varepsilon}_{0}^{B}|\boldsymbol{\mu }]\,\eqsim \Phi \left( 
\frac{-G_{0}^{B,R}+c}{\sqrt{D_{0}^{B,R}}}\right) ,
\end{aligned}
\label{condRaudys}
\end{equation}%
where 
\begin{equation}
G_{0}^{B,R}=G_{0}^{B,f},  \label{bkshbaa}
\end{equation}
with $G_{0}^{B,f}$ being presented in (\ref{g0f}) and 
\begin{equation}
\begin{aligned}
& D_{0}^{B,R}=\delta _{\boldsymbol{\mu }}^{2}+\frac{\delta _{\boldsymbol{\mu 
}}^{2}}{n_{0}(1+\beta _{0})}+\frac{\delta _{\boldsymbol{\mu }}^{2}}{%
n_{1}(1+\beta _{0})}+\frac{\delta _{\boldsymbol{\mu }}^{2}}{n_{0}(1+\beta
_{0})^{2}} \\
& +\frac{\beta _{0}}{(1+\beta _{0})^{2}}\bigg[\frac{\eta _{\mathbf{m}_{0},%
\boldsymbol{\mu }_{1}}-(1-\beta _{0})\eta _{\mathbf{m}_{0},\boldsymbol{\mu }%
_{0}}-\delta _{\boldsymbol{\mu }}^{2}}{n_{0}}+\frac{(1+\beta _{0})\eta _{%
\mathbf{m}_{0},\boldsymbol{\mu }_{1}}-\eta _{\mathbf{m}_{0},\boldsymbol{\mu }%
_{0}}}{n_{1}}\bigg]+\frac{p}{n_{0}}+\frac{p}{n_{1}}+\frac{p}{%
n_{0}^{2}(1+\beta _{0})} \\
& +\frac{p}{n_{0}n_{1}(1+\beta _{0})}+\frac{(1-\beta _{0})^{2}p}{%
2n_{0}^{2}(1+\beta _{0})^{2}}+\frac{p}{n_{0}n_{1}(1+\beta _{0})^{2}}+\frac{p%
}{2n_{1}^{2}}.
\end{aligned}
\label{d0f}
\end{equation}%
The corresponding approximation for $E_{S_{n}}[\hat{\varepsilon}_{1}^{B}|%
\boldsymbol{\mu }]$ is 
\begin{equation}
\begin{aligned}
E_{S_{n}}[\hat{\varepsilon}_{1}^{B}|\boldsymbol{\mu }]\,\eqsim \Phi \left( 
\frac{G_{1}^{B,R}-c}{\sqrt{D_{1}^{B,R}}}\right) ,
\end{aligned}%
\end{equation}%
where $D_{1}^{B,R}$ and $G_{1}^{B,R}$ are obtained by exchanging $n_{0}$ and 
$n_{1}$, $\nu _{0}$ and $\nu _{1}$, $\mathbf{m}_{0}$ and $\mathbf{m}_{1}$,
and $\boldsymbol{\mu }_{0}$ and $\boldsymbol{\mu }_{1}$ in $D_{0}^{B,R}$ and 
$-G_{0}^{B,R}$, respectively. It is straightforward to see that 
\begin{equation}
\begin{aligned}
& G_{0}^{B,R}\;{\overset{K}{\rightarrow }}\;G_{0}^{B}, \\
& D_{0}^{B,R}\;{\overset{K}{\rightarrow }}\;\overline{\delta }_{\boldsymbol{%
\mu }}^{2}+J_{0}+J_{1},
\end{aligned}
\label{bcjlak}
\end{equation}%
with $G_{0}^{B}$ being defined in Theorem 1. Therefore, the approximation
obtained in (\ref{condRaudys}) is asymptotically exact and (\ref%
{eq-tdEe0s2dU}) and (\ref{condRaudys}) are asymptotically equivalent.

\subsection{Unconditional Expectation of\textbf{\ $\hat{\protect\varepsilon}%
_{i}^{B}$:\ $E_{\boldsymbol{\protect\mu },S_{n}}\left[ \hat{\protect%
\varepsilon}_{i}^{B}\right] $}}

We consider the unconditional expectation of $\hat{\varepsilon}_{i}^{B}$
under Bayesian-Kolmogorov asymptotics. The proof of the following theorem is
presented in the Appendix.

\begin{theorem}
\label{thm-m2} Consider the sequence of Gaussian discrimination problems
defined by (\ref{gkusgukll}). Then 
\begin{equation}
\begin{aligned}
\lim_{\substack{ \text{b.k.a.c.}  \\ {}}}E_{\boldsymbol{\mu},S_n}\left[\hat{%
\varepsilon} _{i}^{B}\right]=\Phi \left( (-1)^{i}\;\frac{-H_{i}+c}{\sqrt{F}}%
\right),
\end{aligned}
\label{eq-t1}
\end{equation}%
so that \vspace{-0.2cm} 
\begin{equation}
\lim_{\substack{ \text{b.k.a.c.}  \\ {}}}E_{\boldsymbol{\mu},S_n}\left[\hat{%
\varepsilon} _{i}^{B}\right]\,= {\alpha}_0\Phi \left( \frac{-H_{0}+c}{\sqrt{F%
}}\right) + {\alpha}_1\Phi \left( \frac{H_{1}-c}{\sqrt{F}}\right),
\label{csxasas}
\end{equation}%
where \vspace{-0.2cm} 
\begin{equation}
\begin{aligned}
& H_{0}\,=\,\frac{1}{2}\left(\overline{\Delta}^{2}_{\mathbf{m}}+J_1-J_0+%
\frac{J_0}{\gamma_0}+\frac{J_1}{\gamma_1}\right), \\[-1ex]
&H_{1}\,=\,-\frac{1}{2}\left(\overline{\Delta}^{2}_{\mathbf{m}}+J_0-J_1+%
\frac{J_0}{\gamma_0}+\frac{J_1}{\gamma_1}\right), \\[-1ex]
& F\,=\,\overline{\Delta}^{2}_{\mathbf{m}}+J _{0}+J_{1}+\frac{J_0}{\gamma_0}+%
\frac{J_1}{\gamma_1}\,. \quad \blacksquare
\end{aligned}
\label{njzzp}
\end{equation}
\end{theorem}

Theorem \ref{thm-m2} suggests the finite-sample approximation 
\begin{equation}
\begin{aligned}
E_{\boldsymbol{\mu },S_{n}}\left[ \hat{\varepsilon}_{0}^{B}\right] \eqsim
\Phi \left( \frac{-H_{0}^{R}+c}{\sqrt{\Delta _{\mathbf{m}}^{2}+\frac{p}{n_{0}%
}+\frac{p}{n_{1}}+\frac{p}{\nu _{0}}+\frac{p}{\nu _{1}}}}\right) ,
\end{aligned}
\label{uncondKol}
\end{equation}%
where 
\begin{equation}
H_{0}^{R}=\frac{1}{2}\left( \Delta _{\mathbf{m}}^{2}+\frac{p}{n_{1}}-\frac{p%
}{n_{0}}+\frac{p}{\nu _{0}}+\frac{p}{\nu _{1}}\right) .  \label{jacdbdx}
\end{equation}

From (\ref{rewriteGaussian}) we can get the Raudys-type approximation: 
\begin{equation}
\begin{aligned}
E_{\boldsymbol{\mu },S_{n}}[\hat{\varepsilon}_{0}^{B}]& =E_{\boldsymbol{\mu }%
}\left[ P(U_{0}(\bar{\mathbf{x}}_{0},\bar{\mathbf{x}}_{1},\mathbf{z})\leq
c\,|\mathbf{z}\in \Psi _{0},\boldsymbol{\mu })\right] \eqsim \Phi \left( 
\frac{-E_{\boldsymbol{\mu },S_{n},\mathbf{z}}[U_{0}(\bar{\mathbf{x}}_{0},%
\bar{\mathbf{x}}_{1},\mathbf{z})|\mathbf{z}\in \Psi _{0})]+c}{\sqrt{\text{Var%
}_{\boldsymbol{\mu },S_{n},\mathbf{z}}[U_{0}(\bar{\mathbf{x}}_{0},\bar{%
\mathbf{x}}_{1},\mathbf{z})|\mathbf{z}\in \Psi _{0}]}}\right) .
\end{aligned}%
\end{equation}%
Some algebraic manipulations yield (Suppl. Section B) 
\begin{equation}
\begin{aligned}
E_{\boldsymbol{\mu },S_{n}}[\hat{\varepsilon}_{0}^{B}]\,\eqsim \Phi \left( 
\frac{-H_{0}^{R}+c}{\sqrt{F_{0}^{R}}}\right) ,
\end{aligned}
\label{uncondRaudys}
\end{equation}%
where 
\begin{equation}
\begin{aligned}
& F_{0}^{R}=\left( 1+\frac{1}{\nu _{0}}+\frac{1}{\nu _{1}}+\frac{1}{n_{1}}%
\right) \Delta _{\mathbf{m}}^{2}+p\left( \frac{1}{n_{0}}+\frac{1}{n_{1}}+%
\frac{1}{\nu _{0}}+\frac{1}{\nu _{1}}\right) \\
& +p\left( \frac{1}{2n_{0}^{2}}+\frac{1}{2n_{1}^{2}}+\frac{1}{2\nu _{0}^{2}}+%
\frac{1}{2\nu _{1}^{2}}\right) +p\left( \frac{1}{n_{1}\nu _{0}}+\frac{1}{%
n_{1}\nu _{1}}+\frac{1}{\nu _{0}\nu _{1}}\right) .
\end{aligned}
\label{f0f}
\end{equation}%
It is straightforward to see that 
\begin{equation}
\begin{aligned}
& H_{0}^{R}\;{\overset{K}{\rightarrow }}\;H_{0}, \\
& F_{0}^{R}\;{\overset{K}{\rightarrow }}\;\overline{\Delta }_{\mathbf{m}%
}^{2}+J_{0}+J_{1}+\frac{J_{0}}{\gamma _{0}}+\frac{J_{1}}{\gamma _{1}},
\end{aligned}
\label{ncdlnc}
\end{equation}%
with $H_{0}$ defined in Theorem 2. Hence, the approximation obtained in (\ref%
{uncondRaudys}) is asymptotically exact and both (\ref{uncondKol}) and (\ref%
{uncondRaudys}) are asymptotically equivalent.

\section{Second Moments of $\hat{\protect\varepsilon}_{i}^{B}$}

\label{section5}

Here we employ the Bayesian-Kolmogorov asymptotic analysis to characterize
the second and cross moments with the actual error, and therefore the MSE of
error estimation.

\subsection{Conditional Second and Cross Moments of $\hat{\protect\varepsilon%
}_{i}^{B}$}

Defining two i.i.d. random vectors, $\mathbf{z}$ and $\mathbf{z}^{\prime }$,
yields the second moment representation 
\begin{equation}
\begin{aligned}
& E_{S_{n}}[(\hat{\varepsilon}_{0}^{B})^{2}|\boldsymbol{\mu }%
]=E_{S_{n}}[P(U_{0}(\bar{\mathbf{x}}_{0},\bar{\mathbf{x}}_{1},\mathbf{z}%
)\leq c\,|\bar{\mathbf{x}}_{0},\bar{\mathbf{x}}_{1},\mathbf{z}\in \Psi _{0},%
\boldsymbol{\mu })^{2}] \\
& =E_{S_{n}}\bigg[P(U_{0}(\bar{\mathbf{x}}_{0},\bar{\mathbf{x}}_{1},\mathbf{z%
})\leq c\,|\bar{\mathbf{x}}_{0},\bar{\mathbf{x}}_{1},\mathbf{z}\in \Psi _{0},%
\boldsymbol{\mu })P(U_{0}(\bar{\mathbf{x}}_{0},\bar{\mathbf{x}}_{1},\mathbf{z%
}^{\prime })\leq c\,|\bar{\mathbf{x}}_{0},\bar{\mathbf{x}}_{1},\mathbf{z}%
^{\prime }\in \Psi _{0},\boldsymbol{\mu })\bigg] \\
& =E_{S_{n}}\bigg[P\Big(U_{0}(\bar{\mathbf{x}}_{0},\bar{\mathbf{x}}_{1},%
\mathbf{z})\leq c\,,U_{0}(\bar{\mathbf{x}}_{0},\bar{\mathbf{x}}_{1},\mathbf{z%
}^{\prime })\leq c\,|\bar{\mathbf{x}}_{0},\bar{\mathbf{x}}_{1},\mathbf{z}\in
\Psi _{0},\mathbf{z}^{\prime }\in \Psi _{0},\boldsymbol{\mu }\Big)\bigg] \\
& =P\Big(U_{0}(\bar{\mathbf{x}}_{0},\bar{\mathbf{x}}_{1},\mathbf{z})\leq
c\,,U_{0}(\bar{\mathbf{x}}_{0},\bar{\mathbf{x}}_{1},\mathbf{z}^{\prime
})\leq c\,|\mathbf{z}\in \Psi _{0},\mathbf{z}^{\prime }\in \Psi _{0},%
\boldsymbol{\mu }\Big),
\end{aligned}
\label{kjxbkasU}
\end{equation}%
where $\mathbf{z}$ and $\mathbf{z}^{\prime }$ are independent of $S_{n}$,
and $\Psi _{i}$ is a multivariate Gaussian, $N(\mathbf{m}_{i},\frac{%
(n_{i}+\nu _{i}+1)(n_{i}+\nu _{i})}{\nu _{i}^{2}}\mathbf{\Sigma }) $, and $%
U_{i}(\bar{\mathbf{x}}_{0},\bar{\mathbf{x}}_{1},\mathbf{z})$ being defined
in (\ref{Ui}).%
%
%
%
%
%
%
%
%
%
%
%
%
%
%
%
%
%
%
%
%
%
%
%
%
%
%
%
%
%
%
%
%
%
%
%
%
%
%
%
%
%
%
%
%
%
%
%
%
%
%
%
%
%
%
%
%
%
%

We have the following theorem, with the proof presented in the Appendix.

\begin{theorem}
\label{thm-m4} For the sequence of Gaussian discrimination problems in (\ref%
{gkusguk}) and for $i,j=0,1$, 
\begin{equation}
\begin{aligned}
\lim_{\substack{ \text{b.k.a.c.}  \\ }}E_{S_{n}}[\hat{\varepsilon}_{i}^{B}%
\hat{\varepsilon}_{j}^{B}|\boldsymbol{\mu }]=\Phi \left((-1)^{i}\; \frac{%
-G_{i}^{B}+c}{\sqrt{D}}\right) \Phi \left( (-1)^{j}\;\frac{-G_{j}^{B}+c}{%
\sqrt{D}}\right) ,
\end{aligned}
\label{eq-t1}
\end{equation}%
so that \vspace{-0.2cm} 
\begin{equation}
\lim_{\substack{ \text{b.k.a.c.}  \\ }}E_{S_{n}}[(\hat{\varepsilon}^{B})^{2}|%
\boldsymbol{\mu }]=\left[ {\alpha}_{0}\Phi \left( \frac{-G_{0}^{B}+c}{\sqrt{D%
}}\right) +{\alpha}_{1}\Phi \left( \frac{G_{1}^{B}-c}{\sqrt{D}}\right) %
\right] ^{2},  \label{csxasas}
\end{equation}%
where $G_{0}^{B}$, $G_{1}^{B}$, and $D$ are defined in (\ref{njzz}). \quad $\blacksquare$
\end{theorem}

This theorem suggests the finite-sample approximation 
\begin{equation}
\begin{aligned}
E_{S_{n}}[(\hat{\varepsilon}_{0}^{B})^{2}|\boldsymbol{\mu }]\,\eqsim \left[
\Phi \left( \frac{-G_{0}^{B,f}+c}{\sqrt{\delta _{\boldsymbol{\mu }}^{2}+%
\frac{p}{n_{0}}+\frac{p}{n_{1}}}}\right) \right] ^{2}\,,
\end{aligned}
\label{eq-tdEe0s2dgg}
\end{equation}%
which is the square of the approximation (\ref{eq-tdEe0s2dU}). Corresponding
approximations for $E[\hat{\varepsilon}_{0}^{B}\hat{\varepsilon}_{1}^{B}]$
and $E[(\hat{\varepsilon}_{1}^{B})^{2}]$ are obtained similarly.

Similar to the proof of Theorem \ref{thm-m4}, we obtain the conditional
cross moment of $\hat{\varepsilon}^{B}$.

\begin{theorem}
\label{thm-m5} Consider the sequence of Gaussian discrimination problems in (%
\ref{gkusguk}). Then for $i,j=0,1$, 
\begin{equation}
\begin{aligned}
\lim_{\substack{ \text{b.k.a.c.} \\ }}E_{S_{n}}[\hat{\varepsilon}%
_{i}^{B}\varepsilon _{j}|\boldsymbol{\mu }]=\Phi \left( (-1)^{i}\;\frac{%
-G_{i}^{B}+c}{\sqrt{D}}\right) \Phi \left( (-1)^{j}\;\frac{-G_{j}+c}{\sqrt{D}%
}\right) ,
\end{aligned}
\label{eq-t1bb}
\end{equation}%
so that \vspace{-0.2cm} 
\begin{equation}
\begin{aligned}
\lim_{\substack{ \text{b.k.a.c.} \\ }}E_{S_{n}}[\hat{\varepsilon}%
^{B}\varepsilon |\boldsymbol{\mu }]=\sum_{i=0}^{1}\sum_{j=0}^{1}\bigg[{%
\alpha }_{i}{\alpha }_{j}\Phi \left( (-1)^{i}\;\frac{-G_{i}^{B}+c}{\sqrt{D}}%
\right) \Phi \left( (-1)^{j}\;\frac{-G_{j}+c}{\sqrt{D}}\right) \bigg],
\end{aligned}
\label{csxasasbb}
\end{equation}%
where $G_{i}^{B}$ and $D$ are defined in (\ref{njzz}) and $G_{i}$ is defined
in (\ref{njzzpp}). \quad $\blacksquare$ \vspace{-0.2cm} 
\end{theorem}

This theorem suggests the finite-sample approximation 
\begin{equation}
\begin{aligned}
E_{S_{n}}[\hat{\varepsilon}_{0}^{B}\varepsilon _{0}|\boldsymbol{\mu }]\eqsim
\Phi \left( \frac{-G_{0}^{B,f}+c}{\sqrt{\delta _{\boldsymbol{\mu }}^{2}+%
\frac{p}{n_{0}}+\frac{p}{n_{1}}}}\right) \Phi \left( -\frac{1}{2}\frac{%
\delta _{\boldsymbol{\mu }}^{2}+\frac{p}{n_{1}}-\frac{p}{n_{0}}-c}{\sqrt{%
\delta _{\boldsymbol{\mu }}^{2}+\frac{p}{n_{0}}+\frac{p}{n_{1}}}}\right) .
\end{aligned}
\label{bxjbxk}
\end{equation}%
This is a product of (\ref{eq-tdEe0s2dU}) and the finite-sample
approximation for $E_{S_{n}}[\varepsilon _{0}|\boldsymbol{\mu }]$\ in \cite%
{Zollanvari}.\ 

A consequence of Theorems~\ref{thm-m1},~\ref{thm-m4}, and \ref{thm-m5} is
that all the conditional variances and covariances are asymptotically zero: 
\begin{equation}
\begin{aligned}
\lim_{\substack{ \text{b.k.a.c.}  \\ }}\text{Var}_{S_{n}}(\hat{\varepsilon}%
^{B}|\boldsymbol{\mu })=\lim_{\substack{ \text{b.k.a.c.}  \\ }}\text{Var}%
_{S_{n}}({\varepsilon |\boldsymbol{\mu }})=\lim_{\substack{ \text{b.k.a.c.} 
\\ }}\text{Cov}_{S_{n}}(\varepsilon ,\hat{\varepsilon}^{B}|\boldsymbol{\mu }%
)=0.
\end{aligned}%
\end{equation}%
Hence, the deviation variance is also asymptotically zero, $\lim_{\mathrm{%
b.k.a.c}.}$ $\text{Var}_{S_{n}}^{d}[{\hat{\varepsilon}}^{B}|\boldsymbol{\mu }%
]=0$. Hence, defining the conditional bias as

\begin{equation}
\begin{aligned}
\mathrm{Bias}_{C,n}[\hat{\varepsilon}^{B}]=E_{S_{n}}[\hat{\varepsilon}%
^{B}-\varepsilon |\boldsymbol{\mu }],
\end{aligned}%
\end{equation}%
the asymptotic RMS reduces to%
\begin{equation}
\begin{aligned}
\lim_{\substack{ \text{b.k.a.c.}  \\ }}\text{RMS}_{S_{n}}[{\hat{\varepsilon}}%
^{B}|\boldsymbol{\mu }]=\lim_{\substack{ \text{b.k.a.c.}  \\ }}|\text{Bias}%
_{C,n}[\hat{\varepsilon}^{B}]|.
\end{aligned}
\label{mshhxa}
\end{equation}%
To express the conditional bias, as proven in \cite{Zollanvari}, 
\begin{equation}
\lim_{\substack{ \text{b.k.a.c.}  \\ }}E_{S_{n}}[\varepsilon |\boldsymbol{%
\mu }]\,=\alpha _{0}\Phi \left( \frac{-G_{0}+c}{\sqrt{D}}\right) +\alpha
_{1}\Phi \left( \frac{G_{1}-c}{\sqrt{D}}\right) ,  \label{csxasaspp}
\end{equation}%
where

\begin{equation}
\begin{aligned}
& G_{0}\,=\,\frac{1}{2}(\delta _{\boldsymbol{\mu }}^{2}+J_{1}-J_{0}), \\%
[-1ex]
& G_{1}\,=\,-\frac{1}{2}(\delta _{\boldsymbol{\mu }}^{2}+J_{0}-J_{1}), \\%
[-1ex]
& D\,=\,\delta _{\boldsymbol{\mu }}^{2}+J_{0}+J_{1}\,.
\end{aligned}
\label{njzzpp}
\end{equation}%
It follows from Theorem 1 and (\ref{csxasaspp}) that

\begin{equation}
\begin{aligned}
\!\lim_{\substack{ \text{b.k.a.c.}  \\ }}\text{Bias}_{C,n}[\hat{\varepsilon}%
^{B}]\!=\! \alpha_{0}\!\left[ \Phi \left( \frac{-G_{0}^{B}+c}{\sqrt{D}}%
\right) \!-\!\Phi \left( \frac{-G_{0}+c}{\sqrt{D}}\right) \right] \! +\!
\alpha _{1}\!\left[ \Phi \left( \frac{G_{1}^{B}-c}{\sqrt{D}}\right)
\!-\!\Phi \left( \frac{G_{1}-c}{\sqrt{D}}\right) \right] .
\end{aligned}
\label{condbiasB}
\end{equation}%
Recall that the MMSE error estimator is unconditionally unbiased: $\text{Bias%
}_{U,n}[\hat{\varepsilon}^{B}]\,=\,E_{\boldsymbol{\mu },S_{n}}\left[ \hat{%
\varepsilon}^{B}-\varepsilon \right] =0$.

We next obtain Raudys-type approximations corresponding to Theorems 3 and 4
by utilizing the joint distribution of $U_{i}(\bar{\mathbf{x}}_{0},\bar{%
\mathbf{x}}_{1},\mathbf{z})$ and $U_{j}(\bar{\mathbf{x}}_{0},\bar{\mathbf{x}}%
_{1},\mathbf{z}^{\prime })$, defined in (\ref{Ui}), with $\mathbf{z}$ and $\mathbf{z}^{\prime }$ being independently selected from populations $%
\Psi _{0}$ or $\Psi _{1}$. We employ the function 
\begin{equation}
\Phi (a,b;\rho )=\int\limits_{-\infty }^{a}\int\limits_{-\infty }^{b}\frac{1%
}{2\pi \sqrt{1-\rho ^{2}}}\exp \left\{ \frac{-\left( x^{2}+y^{2}-2\rho
xy\right) }{2(1-\rho ^{2})}\right\} dx\,dy,
\end{equation}%
which is the distribution function of a joint bivariate Gaussian vector with
zero means, unit variances, and correlation coefficient $\rho $. Note that $%
\Phi (a,\infty ;\rho )=\Phi (a)$ and $\Phi (a,b;0)=\Phi (a)\Phi (b)$. For
simplicity of notation, we write $\Phi (a,a;\rho )$ as $\Phi (a;\rho )$. The
rectangular-area probabilities involving any jointly Gaussian pair of
variables $(x,y)$ can be expressed as 
\begin{equation}
P\left( x\leq c,y\leq d\right) \,=\,\Phi \left( \frac{c-\mu _{x}}{\sigma _{x}%
},\frac{d-\mu _{y}}{\sigma _{y}};\,\rho _{xy}\right) ,
\end{equation}%
with $\mu _{x}=E[x]$, $\mu _{y}=E[y]$, $\sigma _{x}=\sqrt{\text{Var}(x)}$, $%
\sigma _{Y}=\sqrt{\text{Var}(y)}$, and correlation coefficient $\rho _{xy}$.

Using (\ref{kjxbkasU}), we obtain the second-order extension of (\ref%
{kshbxkabk}) by 
\begin{equation}
\begin{aligned}
& E_{S_{n}}[(\hat{\varepsilon}_{0}^{B})^{2}|\boldsymbol{\mu }]=P\Big(U_{0}(%
\bar{\mathbf{x}}_{0},\bar{\mathbf{x}}_{1},\mathbf{z})\leq c\,,U_{0}(\bar{%
\mathbf{x}}_{0},\bar{\mathbf{x}}_{1},\mathbf{z}^{\prime })\leq c\,|\mathbf{z}%
\in \Psi _{0},\mathbf{z}^{\prime }\in \Psi _{0},\boldsymbol{\mu }\Big) \\
& \eqsim \Phi \bigg(\frac{-E_{S_{n},\mathbf{z}}[U_{0}(\bar{\mathbf{x}}_{0},%
\bar{\mathbf{x}}_{1},\mathbf{z})\mid \mathbf{z}\in \Psi _{0},\boldsymbol{\mu 
}]+c}{\sqrt{\text{Var}_{S_{n},\mathbf{z}}[U_{0}(\bar{\mathbf{x}}_{0},\bar{%
\mathbf{x}}_{1},\mathbf{z})\mid \mathbf{z}\in \Psi _{0},\boldsymbol{\mu }]}};%
\frac{\text{Cov}_{S_{n},\mathbf{z}}[U_{0}(\bar{\mathbf{x}}_{0},\bar{\mathbf{x%
}}_{1},\mathbf{z}),U_{0}(\bar{\mathbf{x}}_{0},\bar{\mathbf{x}}_{1},\mathbf{z}%
^{\prime })|\mathbf{z}\in \Psi _{0},\mathbf{z}^{\prime }\in \Psi _{0},%
\boldsymbol{\mu }]}{\text{Var}_{S_{n},\mathbf{z}}[U_{0}(\bar{\mathbf{x}}_{0},%
\bar{\mathbf{x}}_{1},\mathbf{z})|\mathbf{z}\in \Psi _{0},\boldsymbol{\mu }]}%
\bigg).
\end{aligned}
\label{eq-2dam112}
\end{equation}%
Using (\ref{eq-2dam112}), some algebraic manipulations yield 
\begin{equation}
\begin{aligned}
E_{S_{n}}[(\hat{\varepsilon}_{0}^{B})^{2}|\boldsymbol{\mu }]\,\eqsim \,\Phi
\left( \frac{-G_{0}^{B,R}+c}{\sqrt{D_{0}^{B,R}}};\frac{C_{0}^{B,R}}{%
D_{0}^{B,R}}\right) ,
\end{aligned}
\label{eq-2dam1123ss}
\end{equation}%
with $G_{0}^{B,R}$ and $D_{0}^{B,R}$ being presented in (\ref{bkshbaa}) and (%
\ref{d0f}), respectively, and 
\begin{equation}
\begin{aligned}
& C_{0}^{B,R}=\text{Cov}_{S_{n},\mathbf{z}}[U_{0}(\bar{\mathbf{x}}_{0},\bar{%
\mathbf{x}}_{1},\mathbf{z}),U_{0}(\bar{\mathbf{x}}_{0},\bar{\mathbf{x}}_{1},%
\mathbf{z}^{\prime })|\mathbf{z}\in \Psi _{0},\mathbf{z}^{\prime }\in \Psi
_{0},\boldsymbol{\mu }] \\
& =\frac{\beta _{0}}{(1+\beta _{0})^{2}}\bigg[\frac{\eta _{\mathbf{m}_{0},%
\boldsymbol{\mu }_{1}}-(1-\beta _{0})\eta _{\mathbf{m}_{0},\boldsymbol{\mu }%
_{0}}-\delta _{\boldsymbol{\mu }}^{2}}{n_{0}}+\frac{(1+\beta _{0})\eta _{%
\mathbf{m}_{0},\boldsymbol{\mu }_{1}}-\eta _{\mathbf{m}_{0},\boldsymbol{\mu }%
_{0}}}{n_{1}}\bigg]+\frac{(1-\beta _{0})^{2}p}{2n_{0}^{2}(1+\beta _{0})^{2}}
\\
& +\frac{p}{n_{0}n_{1}(1+\beta _{0})^{2}}+\frac{p}{2n_{1}^{2}}+\frac{\delta
_{\boldsymbol{\mu }}^{2}}{n_{1}(1+\beta _{0})}+\frac{\delta _{\boldsymbol{%
\mu }}^{2}}{n_{0}(1+\beta _{0})^{2}},
\end{aligned}
\label{c0f}
\end{equation}%
The proof of (\ref{c0f}) follows by expanding $U_{0}(\bar{\mathbf{x}}_{0},%
\bar{\mathbf{x}}_{1},\mathbf{z})$ and $U_{0}(\bar{\mathbf{x}}_{0},\bar{%
\mathbf{x}}_{1},\mathbf{z}^{\prime })$ from (\ref{Ui}) and then using the
set of identities in the proof of (\ref{uncondRaudys}), i.e. equation (S.1)
from Suppl. Section B. Similarly, 
\begin{equation}
\begin{aligned}
E_{S_{n}}[(\hat{\varepsilon}_{1}^{B})^{2}|\boldsymbol{\mu }]=P(U_{1}(\bar{%
\mathbf{x}}_{0},\bar{\mathbf{x}}_{1},\mathbf{z})>c\,,U_{1}(\bar{\mathbf{x}}%
_{0},\bar{\mathbf{x}}_{1},\mathbf{z}^{\prime })>c\,|\mathbf{z}\in \Psi _{1},%
\mathbf{z}^{\prime }\in \Psi _{1},\boldsymbol{\mu })\eqsim \Phi \left( \frac{%
G_{1}^{B,R}-c}{\sqrt{D_{1}^{B,R}}};\frac{C_{1}^{B,R}}{D_{1}^{B,R}}\right) ,
\end{aligned}
\label{eq-2dam1123s}
\end{equation}%
where $D_{1}^{B,R}$, $G_{1}^{B,R}$, and $C_{1}^{B,R}$ are obtained by
exchanging $n_{0}$ and $n_{1}$, $\nu _{0}$ and $\nu _{1}$, $\mathbf{m}_{0}$
and $\mathbf{m}_{1}$, and $\boldsymbol{\mu }_{0}$ and $\boldsymbol{\mu }_{1}$%
, in (\ref{d0f}), in $-G_{0}^{B,f}$ obtained from (\ref{g0f}), and in (\ref%
{c0f}), respectively.

Having $C_{0}^{B,R}\;{\overset{K}{\rightarrow }}\;0$ together with (\ref%
{bcjlak}) shows that (\ref{eq-2dam1123ss}) is asymptotically exact, that is,
asymptotically equivalent to $E_{S_{n}}[(\hat{\varepsilon}_{0}^{B})^{2}|%
\boldsymbol{\mu }]$ obtained in Theorem 3. Similarly, it can be shown that 
\begin{equation}
\begin{aligned}
& E_{S_{n}}[\hat{\varepsilon}_{0}^{B}\hat{\varepsilon}_{1}^{B}|\boldsymbol{%
\mu }]=P(U_{0}(\bar{\mathbf{x}}_{0},\bar{\mathbf{x}}_{1},\mathbf{z})\leq
c\,,-U_{1}(\bar{\mathbf{x}}_{0},\bar{\mathbf{x}}_{1},\mathbf{z}^{\prime
})<-c\,|\mathbf{z}\in \Psi _{0},\mathbf{z}^{\prime }\in \Psi _{1},%
\boldsymbol{\mu }) \\
& \eqsim \Phi \left( \frac{-G_{0}^{B,R}+c}{\sqrt{D_{0}^{B,R}}},\frac{%
G_{1}^{B,R}-c}{\sqrt{D_{1}^{B,R}}};\frac{C_{01}^{B,R}}{\sqrt{%
D_{0}^{B,R}D_{1}^{B,R}}}\right) ,
\end{aligned}
\label{eq-2dam112u}
\end{equation}%
where, after some algebraic manipulations we obtain 
\begin{equation}
\begin{aligned}
& C_{01}^{B,R}=\frac{1}{n_{0}(1+\beta _{0})(1+\beta _{1})}\bigg[\beta
_{0}\eta _{\mathbf{m}_{0},\boldsymbol{\mu }_{0},\boldsymbol{\mu }_{0},%
\boldsymbol{\mu }_{1}}-\beta _{0}\beta _{1}\eta _{\mathbf{m}_{0},\boldsymbol{%
\mu }_{0},\mathbf{m}_{1},\boldsymbol{\mu }_{0}}+\beta _{1}\eta _{\mathbf{m}%
_{1},\boldsymbol{\mu }_{1},\boldsymbol{\mu }_{1},\boldsymbol{\mu }%
_{0}}+\beta _{1}\delta _{\boldsymbol{\mu }}^{2}+\delta _{\boldsymbol{\mu }%
}^{2}\bigg] \\
& +\frac{1}{n_{1}(1+\beta _{0})(1+\beta _{1})}\bigg[\beta _{1}\eta _{\mathbf{%
m}_{1},\boldsymbol{\mu }_{1},\boldsymbol{\mu }_{1},\boldsymbol{\mu }%
_{0}}-\beta _{0}\beta _{1}\eta _{\mathbf{m}_{1},\boldsymbol{\mu }_{1},%
\mathbf{m}_{0},\boldsymbol{\mu }_{1}}+\beta _{0}\eta _{\mathbf{m}_{0},%
\boldsymbol{\mu }_{0},\boldsymbol{\mu }_{0},\boldsymbol{\mu }_{1}}+\beta
_{0}\delta _{\boldsymbol{\mu }}^{2}+\delta _{\boldsymbol{\mu }}^{2}\bigg] \\
& +\frac{p}{n_{0}n_{1}(1+\beta _{0})(1+\beta _{1})}+\frac{(1-\beta _{0})p}{%
2n_{0}^{2}(1+\beta _{0})}+\frac{(1-\beta _{1})p}{2n_{1}^{2}(1+\beta _{1})}.
\end{aligned}
\label{c01f}
\end{equation}%
Suppl. Section C gives the proof of (\ref{c01f}). Since $C_{01}^{B,R}\;{%
\overset{K}{\rightarrow }}\;0$, (\ref{eq-2dam112u}) is asymptotically exact,
i.e. (\ref{eq-2dam112u}) becomes equivalent to the result of Theorem 3. We
obtain the conditional cross moment similarly: 
\begin{equation}
\begin{aligned}
& E_{S_{n}}[\hat{\varepsilon}_{0}^{B}\varepsilon _{0}|\boldsymbol{\mu }%
]=P(U_{0}(\bar{\mathbf{x}}_{0},\bar{\mathbf{x}}_{1},\mathbf{z})\leq c\,,W(%
\bar{\mathbf{x}}_{0},\bar{\mathbf{x}}_{1},\mathbf{x})\leq c\,|\mathbf{z}\in
\Psi _{0},\mathbf{x}\in \Pi _{0},\boldsymbol{\mu }) \\
& \eqsim \Phi \Bigg(\frac{-E_{S_{n},\mathbf{z}}[U_{0}(\bar{\mathbf{x}}_{0},%
\bar{\mathbf{x}}_{1},\mathbf{z})|\mathbf{z}\in \Psi _{0},\boldsymbol{\mu }]+c%
}{\sqrt{\text{Var}_{S_{n},\mathbf{z}}[U_{0}(\bar{\mathbf{x}}_{0},\bar{%
\mathbf{x}}_{1},\mathbf{z})|\mathbf{z}\in \Psi _{0},\boldsymbol{\mu }]}},%
\frac{-E_{S_{n},\mathbf{x}}[W(\bar{\mathbf{x}}_{0},\bar{\mathbf{x}}_{1},%
\mathbf{x})|\mathbf{x}\in \Pi _{0},\boldsymbol{\mu }]+c}{\sqrt{\text{Var}%
_{S_{n},\mathbf{x}}[W(\bar{\mathbf{x}}_{0},\bar{\mathbf{x}}_{1},\mathbf{x})|%
\mathbf{x}\in \Pi _{0},\boldsymbol{\mu }]}}; \\
& \frac{\text{Cov}_{S_{n},\mathbf{z},\mathbf{x}}[U_{0}(\bar{\mathbf{x}}_{0},%
\bar{\mathbf{x}}_{1},\mathbf{z}),W(\bar{\mathbf{x}}_{0},\bar{\mathbf{x}}_{1},%
\mathbf{x})|\mathbf{z}\in \Psi _{0},\mathbf{x}\in \Pi _{0},\boldsymbol{\mu }]%
}{\sqrt{V_{U_{0}}^{\text{C}}V_{W}^{\text{C}}}}\Bigg),
\end{aligned}
\label{eq-2dam112tr}
\end{equation}%
where 
\begin{equation}
\begin{aligned}
& V_{U_{0}}^{\text{C}}=\text{Var}_{S_{n},\mathbf{z}}[U_{0}(\bar{\mathbf{x}}%
_{0},\bar{\mathbf{x}}_{1},\mathbf{z})|\mathbf{z}\in \Psi _{0},\boldsymbol{%
\mu }], \\
& V_{W}^{\text{C}}=\text{Var}_{S_{n},\mathbf{x}}[W(\bar{\mathbf{x}}_{0},\bar{%
\mathbf{x}}_{1},\mathbf{x})|\mathbf{x}\in \Pi _{0},\boldsymbol{\mu }],
\end{aligned}%
\end{equation}%
where superscript ``$\text{C"}$ denotes conditional variance. Algebraic
manipulations like those leading to (\ref{c0f}) yield 
\begin{equation}
\begin{aligned}
E_{S_{n}}[\hat{\varepsilon}_{0}^{B}\varepsilon _{0}|\boldsymbol{\mu }]\eqsim
\Phi \left( \frac{-G_{0}^{B,R}+c}{\sqrt{D_{0}^{B,R}}},\frac{-G_{0}^{R}+c}{%
\sqrt{D_{0}^{R}}};\frac{C_{0}^{BT,R}}{\sqrt{D_{0}^{B,R}D_{0}^{R}}}\right) ,
\end{aligned}
\label{eq-2dam112gb}
\end{equation}%
where 
\begin{equation}
\begin{aligned}
C_{0}^{BT,R}=\frac{1}{n_{1}(1+\beta _{0})}\bigg[\delta _{\boldsymbol{\mu }%
}^{2}+\beta _{0}\delta _{\boldsymbol{\mu }}^{2}+\beta _{0}\eta _{\mathbf{m}%
_{0},\boldsymbol{\mu }_{0},\boldsymbol{\mu }_{0},\boldsymbol{\mu }_{1}}\bigg]%
-\frac{(1-\beta _{0})p}{2n_{0}^{2}(1+\beta _{0})}+\frac{p}{2n_{1}^{2}},
\end{aligned}%
\end{equation}%
and $G_{0}^{R}$ and $D_{0}^{R}$ having been obtained previously in equations
(49) and (50) of \cite{Zollanvari}, namely, 
\begin{equation}
\begin{aligned}
& G_{0}^{R}=E_{S_{n},\mathbf{x}}[W(\bar{\mathbf{x}}_{0},\bar{\mathbf{x}}_{1},%
\mathbf{x})\mid \mathbf{x}\in \Pi _{0},\boldsymbol{\mu }]=\frac{1}{2}\left(
\delta _{\boldsymbol{\mu }}^{2}+\frac{p}{n_{1}}-\frac{p}{n_{0}}\right) , \\
& D_{0}^{R}=\text{Var}_{S_{n},\mathbf{z}}[W(\bar{\mathbf{x}}_{0},\bar{%
\mathbf{x}}_{1},\mathbf{x})\mid \mathbf{x}\in \Pi _{0},\boldsymbol{\mu }%
]=\delta _{\boldsymbol{\mu }}^{2}+\frac{\delta _{\boldsymbol{\mu }}^{2}}{%
n_{1}}+p\left( \frac{1}{n_{0}}+\frac{1}{n_{1}}+\frac{1}{2n_{0}^{2}}+\frac{1}{%
2n_{1}^{2}}\right) .
\end{aligned}
\label{bjsbkjas}
\end{equation}%
Similarly, we can show that 
\begin{equation}
\begin{aligned}
E_{S_{n}}[\hat{\varepsilon}_{1}^{B}\varepsilon _{1}|\boldsymbol{\mu }]\eqsim
\Phi \left( \frac{G_{1}^{B,R}-c}{\sqrt{D_{1}^{B,R}}},\frac{G_{1}^{R}-c}{%
\sqrt{D_{1}^{R}}};\frac{C_{1}^{BT,R}}{\sqrt{D_{1}^{B,R}D_{1}^{R}}}\right) ,
\end{aligned}
\label{eq-2dam112x}
\end{equation}%
where $D_{1}^{B,R}$ and $G_{1}^{B,R}$ are obtained as in (\ref{eq-2dam1123s}%
), and $D_{1}^{R}$, $G_{1}^{R}$, and $C_{1}^{BT,R}$ are obtained by
exchanging $n_{0}$ and $n_{1}$ in $D_{0}^{R}$, $-G_{0}^{R}$, and $%
C_{0}^{BT,R}$, respectively. Similarly, 
\begin{equation}
\begin{aligned}
E_{S_{n}}[\hat{\varepsilon}_{0}^{B}\varepsilon _{1}|\boldsymbol{\mu }]\eqsim
\Phi \left( \frac{-G_{0}^{B,R}+c}{\sqrt{D_{0}^{B,R}}},\frac{G_{1}^{R}-c}{%
\sqrt{D_{1}^{R}}};\frac{C_{01}^{BT,R}}{\sqrt{D_{0}^{B,R}D_{1}^{R}}}\right) ,
\end{aligned}
\label{eq-2dam112gcb}
\end{equation}%
where 
\begin{equation}
\begin{aligned}
C_{01}^{BT,R}=\frac{1}{n_{0}(1+\beta _{0})}\bigg[\delta _{\boldsymbol{\mu }%
}^{2}+\beta _{0}\eta _{\mathbf{m}_{0},\boldsymbol{\mu }_{0},\boldsymbol{\mu }%
_{0},\boldsymbol{\mu }_{1}}\bigg]+\frac{(1-\beta _{0})p}{2n_{0}^{2}(1+\beta
_{0})}-\frac{p}{2n_{1}^{2}},
\end{aligned}%
\end{equation}%
and 
\begin{equation}
\begin{aligned}
E_{S_{n}}[\hat{\varepsilon}_{1}^{B}\varepsilon _{0}|\boldsymbol{\mu }]\eqsim
\Phi \left( \frac{G_{1}^{B,R}-c}{\sqrt{D_{1}^{B,R}}},\frac{-G_{0}^{R}+c}{%
\sqrt{D_{0}^{R}}};\frac{C_{10}^{BT,R}}{\sqrt{D_{1}^{B,R}D_{0}^{R}}}\right) ,
\end{aligned}%
\end{equation}%
where $C_{10}^{BT,R}$ is obtained by exchanging $n_{0}$ and $n_{1}$, $\nu
_{0}$ and $\nu _{1}$, $\mathbf{m}_{0}$ and $\mathbf{m}_{1}$, and $%
\boldsymbol{\mu }_{0}$ and $\boldsymbol{\mu }_{1}$ in $C_{01}^{BT,R}$.

We see that $C_{0}^{BT,R}\;{\overset{K}{\rightarrow }}\;0$, $C_{1}^{BT,R}\;{%
\overset{K}{\rightarrow }}\;0$, and $C_{01}^{BT,R}\;{\overset{K}{\rightarrow 
}}\;0$. Therefore, from (\ref{bcjlak}) and the fact that $G_{0}^{R}\;{%
\overset{K}{\rightarrow }}\;\overline{\delta }_{\boldsymbol{\mu }%
}^{2}+J_{1}-J_{0}$ and $D_{0}^{R}\;{\overset{K}{\rightarrow }}\;\overline{%
\delta }_{\boldsymbol{\mu }}^{2}+J_{0}+J_{1}$, we see that expressions (\ref%
{eq-2dam112gb}), (\ref{eq-2dam112x}), and (\ref{eq-2dam112gcb}), are all
asymptotically exact (compare to Theorem 4).

\subsection{Unconditional Second and Cross Moments of\textbf{\ $\hat{\protect%
\varepsilon}_{i}^{B}$}}

Similarly to the way (\ref{kjxbkasU}) was obtained, we can show that 
\begin{equation}
\begin{aligned}
& E_{\boldsymbol{\mu},{S_{n}}}[(\hat{\varepsilon}_{0}^{B})^{2}] =E_{%
\boldsymbol{\mu},{S_{n}}}\big[P(U_{0}(\bar{\mathbf{x}}_{0},\bar{\mathbf{x}}%
_{1},\mathbf{z})\leq c\,|\bar{\mathbf{x}}_{0},\bar{\mathbf{x}}_{1},\mathbf{z}%
\in \Psi _{0},\boldsymbol{\mu})^{2}\big] \\
& =E_{\boldsymbol{\mu},{S_{n}}}\bigg[P\Big(U_{0}(\bar{\mathbf{x}}_{0},\bar{%
\mathbf{x}}_{1},\mathbf{z})\leq c\,,U_{0}(\bar{\mathbf{x}}_{0},\bar{\mathbf{x%
}}_{1},\mathbf{z}^{\prime })\leq c\,|\bar{\mathbf{x}}_{0},\bar{\mathbf{x}}%
_{1}, \mathbf{z}\in \Psi _{0},\mathbf{z}^{\prime }\in \Psi _{0},\boldsymbol{%
\mu}\Big)\bigg] \\
& =P\Big(U_{0}(\bar{\mathbf{x}}_{0},\bar{\mathbf{x}}_{1},\mathbf{z})\leq
c\,,U_{0}(\bar{\mathbf{x}}_{0},\bar{\mathbf{x}}_{1},\mathbf{z}^{\prime
})\leq c\,|\mathbf{z}\in \Psi _{0},\mathbf{z}^{\prime }\in \Psi _{0}\Big).
\end{aligned}%
\end{equation}

Similarly to the proofs of Theorem \ref{thm-m4} and \ref{thm-m5}, we get the
following theorems.

\begin{theorem}
\label{thm-m6} Consider the sequence of Gaussian discrimination problems in (%
\ref{gkusgukll}). For $i,j=0,1$, 
\begin{equation}
\begin{aligned}
\lim_{\substack{ \text{b.k.a.c.}  \\ }}E_{\boldsymbol{\mu },S_{n}}[\hat{%
\varepsilon}_{i}^{B}\hat{\varepsilon}_{j}^{B}]=\Phi \left( (-1)^{i}\,\frac{%
-H_{i}+c}{\sqrt{F}}\right) \Phi \left( (-1)^{j}\,\frac{-H_{j}+c}{\sqrt{F}}%
\right) ,
\end{aligned}
\label{eq-t1gg}
\end{equation}%
so that \vspace{-0.2cm} 
\begin{equation}
\lim_{\substack{ \text{b.k.a.c.}  \\ }}E_{\boldsymbol{\mu },S_{n}}[(\hat{%
\varepsilon}^{B})^{2}]\,=\left[ {\alpha}_{0}\Phi \left( \frac{-H_{0}+c}{%
\sqrt{F}}\right) +{\alpha}_{1}\Phi \left( \frac{H_{1}-c}{\sqrt{F}}\right) %
\right] ^{2},  \label{csxasasgg}
\end{equation}%
where $H_{0}$, $H_{1}$, and $F$ are defined in (\ref{njzzp}). \quad $\blacksquare$
\end{theorem}

\begin{theorem}
\label{thm-m7} Consider the sequence of Gaussian discrimination problems in (%
\ref{gkusgukll}). For $i,j=0,1$, 
\begin{equation}
\begin{aligned}
\lim_{\substack{ \text{b.k.a.c.}  \\ }}E_{\boldsymbol{\mu },S_{n}}[\hat{%
\varepsilon}_{i}^{B}\varepsilon _{j}]=\lim_{\substack{ \text{b.k.a.c.}  \\ }}%
E_{\boldsymbol{\mu },S_{n}}[\hat{\varepsilon}_{i}^{B}\hat{\varepsilon}%
_{j}^{B}]=\lim_{\substack{ \text{b.k.a.c.}  \\ }}E_{\boldsymbol{\mu }%
,S_{n}}[\varepsilon _{i}\varepsilon _{j}],
\end{aligned}
\label{eq-t1bbhh}
\end{equation}%
so that \vspace{-0.2cm} 
\begin{equation}
\begin{aligned}
\lim_{\substack{ \text{b.k.a.c.}  \\ }}E_{\boldsymbol{\mu },S_{n}}[\hat{%
\varepsilon}^{B}\varepsilon ]=\sum_{i=0}\sum_{j=0}\bigg[{\alpha }_{i}{\alpha}%
_{j}\Phi \left( (-1)^{i}\,\frac{H_{i}+c}{\sqrt{F}}\right) \Phi \left(
(-1)^{j}\,\frac{H_{j}+c}{\sqrt{F}}\right) \bigg],
\end{aligned}
\label{csxasasbbhh}
\end{equation}%
where $H_{0}$, $H_{1}$, and $F$ are defined in (\ref{njzzp}). \quad $\blacksquare$
\end{theorem}

Theorems \ref{thm-m6} and \ref{thm-m7} suggest the finite-sample
approximation: 
\begin{equation}
\begin{aligned}
E_{\boldsymbol{\mu },S_{n}}[\hat{\varepsilon}_{0}^{B}\hat{\varepsilon}%
_{0}^{B}]\eqsim E_{\boldsymbol{\mu },S_{n}}[\hat{\varepsilon}%
_{0}^{B}\varepsilon _{0}]\eqsim E_{\boldsymbol{\mu },S_{n}}[\varepsilon
_{0}\varepsilon _{0}]\eqsim \left[ \Phi \left( -\frac{1}{2}\frac{ \Delta _{%
\mathbf{m}}^{2}+\frac{p}{n_{1}}-\frac{p}{n_{0}}+\frac{p}{\nu _{0}}+\frac{p}{%
\nu _{1}}-c}{\sqrt{\Delta _{\mathbf{m}}^{2}+\frac{p}{n_{0}}+\frac{p}{n_{1}}+%
\frac{p}{\nu _{0}}+\frac{p}{\nu _{1}}}}\right) \right] ^{2}.
\end{aligned}
\label{ksksksas}
\end{equation}

A consequence of Theorems~\ref{thm-m2},~\ref{thm-m6}, and \ref{thm-m7} is
that 
\begin{equation}
\begin{aligned}
& \lim_{\substack{ \text{b.k.a.c.}  \\ }}\text{Var}_{\boldsymbol{\mu }%
,S_{n}}^{d}[{\hat{\varepsilon}}^{B}]=\lim_{\substack{ \text{b.k.a.c.}  \\ }}|%
\text{Bias}_{U,n}[{\hat{\varepsilon}}^{B}]|=\lim_{\substack{ \text{b.k.a.c.} 
\\ }}\text{Var}_{\boldsymbol{\mu },S_{n}}(\hat{\varepsilon}^{B})=\lim 
_{\substack{ \text{b.k.a.c.}  \\ }}\text{Var}_{\boldsymbol{\mu },S_{n}}({%
\varepsilon }) \\
& =\lim_{\substack{ \text{b.k.a.c.}  \\ }}\text{Cov}_{\boldsymbol{\mu }%
,S_{n}}(\varepsilon ,\hat{\varepsilon}^{B})=\lim_{\substack{ \text{b.k.a.c.} 
\\ }}\text{RMS}_{\boldsymbol{\mu },S_{n}}[\hat{\varepsilon}^{B}]=0.
\end{aligned}
\label{unRMS}
\end{equation}%
\qquad

In \cite{Lori3}, it was shown that $\hat{\varepsilon}^{B}$ is strongly
consistent, meaning that $\hat{\varepsilon}^{B}(S_{n})-\varepsilon
(S_{n})\rightarrow 0$ almost surely as $n\rightarrow \infty $ under rather
general conditions, in particular, for the Gaussian and discrete models
considered in that paper. It was also shown that $\text{MSE}_{\boldsymbol{%
\mu }}[\hat{\varepsilon}^{B}|S_{n}]\rightarrow 0$ almost surely as $%
n\rightarrow \infty $ under similar conditions. Here, we have shown that $%
\text{MSE}_{\boldsymbol{\mu },S_{n}}[\hat{\varepsilon}^{B}]{\overset{K}{%
\rightarrow }}0$ under conditions stated in (\ref{KACL2U}). Some researchers
refer to conditions of double asymptoticity as \textquotedblleft comparable"
dimensionality and sample size \cite{Deev:70,Serd:00}. Therefore, one may
think of $\text{MSE}_{\boldsymbol{\mu },S_{n}}[\hat{\varepsilon}^{B}]{%
\overset{K}{\rightarrow }}0$ meaning that $\text{MSE}_{\boldsymbol{\mu }%
,S_{n}}[\hat{\varepsilon}^{B}]$ is close to zero for asymptotic and
comparable dimensionality, sample size, and certainty parameter.

We now consider Raudys-type approximations. Analogous to the approximation
used in (\ref{eq-2dam112}), we obtain the unconditional second moment of $%
\hat{\varepsilon}_{0}^{B}$: 
\begin{equation}
\begin{aligned}
& E_{\boldsymbol{\mu },S_{n}}[(\hat{\varepsilon}_{0}^{B})^{2}]\eqsim \Phi %
\Bigg(\frac{-E_{\boldsymbol{\mu },S_{n},\mathbf{z}}[U_{0}(\bar{\mathbf{x}}%
_{0},\bar{\mathbf{x}}_{1},\mathbf{z})\mid \mathbf{z}\in \Psi _{0}]+c}{\sqrt{%
\text{Var}_{\boldsymbol{\mu },S_{n},\mathbf{z}}[U_{0}(\bar{\mathbf{x}}_{0},%
\bar{\mathbf{x}}_{1},\mathbf{z})\mid \mathbf{z}\in \Psi _{0}]}}; \\
& \frac{\text{Cov}_{\boldsymbol{\mu },S_{n},\mathbf{z}}[U_{0}(\bar{\mathbf{x}%
}_{0},\bar{\mathbf{x}}_{1},\mathbf{z}),U_{0}(\bar{\mathbf{x}}_{0},\bar{%
\mathbf{x}}_{1},\mathbf{z}^{\prime })|\mathbf{z}\in \Psi _{0},\mathbf{z}%
^{\prime }\in \Psi _{0})]}{\text{Var}_{\boldsymbol{\mu },S_{n},\mathbf{z}%
}[U_{0}(\bar{\mathbf{x}}_{0},\bar{\mathbf{x}}_{1},\mathbf{z})|\mathbf{z}\in
\Psi _{0}]}\Bigg).
\end{aligned}
\label{eq-2dam112bc}
\end{equation}%
Using (\ref{eq-2dam112bc}) we get

\begin{equation}
\begin{aligned}
E_{\boldsymbol{\mu },S_{n}}[(\hat{\varepsilon}_{0}^{B})^{2}]\,=\,\Phi \left( 
\frac{-H_{0}^{R}+c}{\sqrt{F_{0}^{R}}};\frac{K_{0}^{B,R}}{F_{0}^{R}}\right) \,
\end{aligned}
\label{eq-2dam1123sskk}
\end{equation}%
with $H_{0}^{R}$ and $F_{0}^{R}$ given in (\ref{jacdbdx}) and (\ref{f0f}),
respectively, and 
\begin{equation}
\begin{aligned}
& K_{0}^{B,R}=\text{Cov}_{\boldsymbol{\mu },S_{n},\mathbf{z}}[U_{0}(\bar{%
\mathbf{x}}_{0},\bar{\mathbf{x}}_{1},\mathbf{z}),U_{0}(\bar{\mathbf{x}}_{0},%
\bar{\mathbf{x}}_{1},\mathbf{z}^{\prime })|\mathbf{z}\in \Psi _{0},\mathbf{z}%
^{\prime }\in \Psi _{0})] \\
& =\left( \frac{1}{n_{0}(1+\beta _{0})^{2}}+\frac{1}{n_{1}}+\frac{1}{\nu
_{0}(1+\beta _{0})^{2}}+\frac{1}{\nu _{1}}\right) \Delta _{\mathbf{m}}^{2}+%
\frac{p}{2n_{0}^{2}} \\
& +\frac{p}{2\nu _{0}^{2}}-\frac{p}{n_{0}\nu _{0}}+\frac{p}{n_{1}\nu _{1}}+%
\frac{p}{2n_{1}^{2}}+\frac{p}{2\nu _{1}^{2}}+\frac{p}{n_{0}n_{1}(1+\beta
_{0})^{2}}+\frac{p}{n_{0}\nu _{1}(1+\beta _{0})^{2}}+\frac{p}{n_{1}\nu
_{0}(1+\beta _{0})^{2}},
\end{aligned}
\label{k0f}
\end{equation}%
Suppl. Section D presents the proof of (\ref{k0f}). In a similar way, 
\begin{equation}
\begin{aligned}
E_{\boldsymbol{\mu },S_{n}}[(\hat{\varepsilon}_{1}^{B})^{2}]\,=\,\Phi \left( 
\frac{H_{1}^{R}-c}{\sqrt{F_{1}^{R}}};\frac{K_{1}^{B,R}}{F_{1}^{R}}\right) ,
\end{aligned}
\label{eq-2dam1123sshh}
\end{equation}%
where $F_{1}^{R}$, $H_{1}^{R}$, and $K_{1}^{B,R}$ are obtained by exchanging 
$n_{0}$ and $n_{1}$, $\nu _{0}$ and $\nu _{1}$, $\mathbf{m}_{0}$ and $%
\mathbf{m}_{1}$, and $\boldsymbol{\mu }_{0}$ and $\boldsymbol{\mu }_{1}$, in
(\ref{f0f}), in $-H_{0}^{B,f}$ obtained from (\ref{jacdbdx}), and (\ref{k0f}%
), respectively.

Having $K_{0}^{B,R}\;{\overset{K}{\rightarrow }}\;0$ together with (\ref%
{ncdlnc}) makes (\ref{eq-2dam1123sskk}) asymptotically exact. We similarly
obtain 
\begin{equation}
\begin{aligned}
E_{\boldsymbol{\mu },S_{n}}[\hat{\varepsilon}_{0}^{B}\hat{\varepsilon}%
_{1}^{B}]=\Phi \left( \frac{-H_{0}^{R}+c}{\sqrt{F_{0}^{R}}},\frac{H_{1}^{R}-c%
}{\sqrt{F_{1}^{R}}};\frac{K_{01}^{B,R}}{\sqrt{F_{0}^{R}F_{1}^{R}}}\right) ,
\end{aligned}
\label{eq-2dam112oo}
\end{equation}%
where 
\begin{equation}
\begin{aligned}
& K_{01}^{B,R}=\frac{p}{(n_{0}+\nu _{0})(n_{1}+\nu _{1})}+\frac{(n_{0}-\nu
_{0})p}{2n_{0}^{2}(n_{0}+\nu _{0})}+\frac{(n_{1}-\nu _{1})p}{%
2n_{1}^{2}(n_{1}+\nu _{1})} \\
& +\frac{n_{0}n_{1}p}{\nu _{0}\nu _{1}(n_{0}+\nu _{0})(n_{1}+\nu _{1})}+%
\frac{(n_{0}-\nu _{0})p}{2\nu _{0}^{2}(n_{0}+\nu _{0})}+\frac{(n_{1}-\nu
_{1})p}{2\nu _{1}^{2}(n_{1}+\nu _{1})} \\
& +\frac{1}{n_{0}+\nu _{0}}\left( 1+\frac{n_{0}}{n_{1}+\nu _{1}}-\frac{\nu
_{0}}{n_{0}}\right) \frac{p}{\nu _{0}}+\frac{1}{n_{1}+\nu _{1}}\left( 1+%
\frac{n_{1}}{n_{0}+\nu _{0}}-\frac{\nu _{1}}{n_{1}}\right) \frac{p}{\nu _{1}}%
+\left( \frac{1}{\nu _{0}}+\frac{1}{\nu _{1}}\right) \Delta _{\mathbf{m}%
}^{2}.
\end{aligned}
\label{k01f}
\end{equation}%
Suppl. Section E presents the proof of (\ref{k01f}). Since $K_{01}^{B,R}\;{%
\overset{K}{\rightarrow }}\;0$, (\ref{eq-2dam112oo}) is asymptotically exact
(compare to Theorem 5). 
Similar to (\ref{eq-2dam112tr}) and (\ref{eq-2dam112gb}), where we
characterized conditional cross moments, we can get the unconditional cross
moments as follows: 
\begin{equation}
\begin{aligned}
& E_{\boldsymbol{\mu },S_{n}}[\hat{\varepsilon}_{0}^{B}\varepsilon
_{0}]\,=P(U_{0}(\bar{\mathbf{x}}_{0},\bar{\mathbf{x}}_{1},\mathbf{z})\leq
c\,,W(\bar{\mathbf{x}}_{0},\bar{\mathbf{x}}_{1},\mathbf{x})\leq c\,\mid 
\mathbf{z}\in \Psi _{0},\mathbf{x}\in \Pi _{0}) \\
& \eqsim \Phi \Bigg(\frac{-E_{\boldsymbol{\mu },S_{n},\mathbf{z}}[U_{0}(\bar{%
\mathbf{x}}_{0},\bar{\mathbf{x}}_{1},\mathbf{z})\mid \mathbf{z}\in \Psi
_{0}]+c}{\sqrt{\text{Var}_{\boldsymbol{\mu },S_{n},\mathbf{z}}[U_{0}(\bar{%
\mathbf{x}}_{0},\bar{\mathbf{x}}_{1},\mathbf{z})\mid \mathbf{z}\in \Psi _{0}]%
}},\frac{-E_{\boldsymbol{\mu },S_{n},\mathbf{x}}[W(\bar{\mathbf{x}}_{0},\bar{%
\mathbf{x}}_{1},\mathbf{x})\mid \mathbf{x}\in \Pi _{0}]+c}{\sqrt{\text{Var}_{%
\boldsymbol{\mu },S_{n},\mathbf{x}}[W(\bar{\mathbf{x}}_{0},\bar{\mathbf{x}}%
_{1},\mathbf{x})\mid \mathbf{x}\in \Pi _{0}]}}; \\
& \frac{\text{Cov}_{\boldsymbol{\mu },S_{n},\mathbf{z},\mathbf{x}}[U_{0}(%
\bar{\mathbf{x}}_{0},\bar{\mathbf{x}}_{1},\mathbf{z}),W(\bar{\mathbf{x}}_{0},%
\bar{\mathbf{x}}_{1},\mathbf{x})|\mathbf{z}\in \Psi _{0},\mathbf{x}\in \Pi
_{0}]}{\sqrt{V_{U_{0}}^{U}V_{W}^{U}}}\Bigg)=\Phi \left( \frac{-H_{0}^{R}+c}{%
\sqrt{F_{0}^{R}}};\frac{K_{0}^{BT,R}}{F_{0}^{R}}\right) ,
\end{aligned}
\label{eq-2dam112trss}
\end{equation}%
where 
\begin{equation}
\begin{aligned}
& V_{U_{0}}^{\text{U}}={\text{Var}_{\boldsymbol{\mu },S_{n},\mathbf{z}%
}[U_{0}(\bar{\mathbf{x}}_{0},\bar{\mathbf{x}}_{1},\mathbf{z})\mid \mathbf{z}%
\in \Psi _{0}]}, \\
& V_{W}^{\text{U}}={\text{Var}_{\boldsymbol{\mu },S_{n},\mathbf{x}}[W(\bar{%
\mathbf{x}}_{0},\bar{\mathbf{x}}_{1},\mathbf{x})\mid \mathbf{x}\in \Pi _{0}]}%
,
\end{aligned}%
\end{equation}%
the superscript ``$\text{U"}$ representing the unconditional variance, $%
H_{0}^{R}$ and $F_{0}^{R}$ being presented in (\ref{jacdbdx}) and (\ref{f0f}%
), respectively, and 
\begin{equation}
\begin{aligned}
K_{0}^{BT,R}& =\left( \frac{n_{0}}{\nu _{0}(n_{0}+\nu _{0})}+\frac{1}{n_{1}}+%
\frac{1}{\nu _{1}}\right) \Delta _{\mathbf{m}}^{2}+\frac{p}{2n_{1}^{2}}+%
\frac{p}{2\nu _{1}^{2}}+\frac{p}{n_{1}\nu _{1}} \\
& +\frac{n_{0}p}{n_{1}\nu _{0}(n_{0}+\nu _{0})}-\frac{(n_{0}-\nu _{0})p}{%
2n_{0}^{2}(n_{0}+\nu _{0})}+\frac{(n_{0}-\nu _{0})p}{2\nu _{0}^{2}(n_{0}+\nu
_{0})}+\frac{n_{0}p}{\nu _{0}\nu _{1}(n_{0}+\nu _{0})}.
\end{aligned}
\label{k0Tf}
\end{equation}%
The proof of (\ref{k0Tf}) is presented in Suppl. Section F. Similarly, 
\begin{equation}
\begin{aligned}
E_{\boldsymbol{\mu },S_{n}}[\hat{\varepsilon}_{0}^{B}\varepsilon _{1}]\eqsim
\Phi \left( \frac{-H_{0}^{R}+c}{\sqrt{F_{0}^{R}}},\frac{H_{1}^{R}-c}{\sqrt{%
F_{1}^{R}}};\frac{K_{01}^{BT,R}}{\sqrt{F_{0}^{R}F_{1}^{R}}}\right) \,
\end{aligned}
\label{eq-2dam112trsso}
\end{equation}%
where, 
\begin{equation}
\begin{aligned}
K_{01}^{BT,R}=\left( \frac{1}{\nu _{0}}+\frac{1}{\nu _{1}}\right) \Delta _{%
\mathbf{m}}^{2}+\frac{p}{2\nu _{0}^{2}}+\frac{p}{2\nu _{1}^{2}}+\frac{p}{\nu
_{0}\nu _{1}}-\frac{p}{2n_{0}^{2}}-\frac{p}{2n_{1}^{2}}.
\end{aligned}
\label{k01Tf}
\end{equation}%
See Suppl. Section G for the proof of (\ref{k01Tf}). Having $K_{0}^{BT,R}\;{%
\overset{K}{\rightarrow }}\;0$ and $K_{01}^{BT,R}\;{\overset{K}{\rightarrow }%
}\;0$ along with (\ref{ncdlnc}) makes (\ref{eq-2dam112trss}) and (\ref%
{eq-2dam112trsso}) asymptotically exact (compare to Theorem 6).

\subsection{Conditional and Unconditional Second Moment of\textbf{\ $\protect%
\varepsilon _{i}$}}

To complete the derivations and obtain the unconditional $\text{RMS}$ of
estimation, we need the conditional and unconditional second moment of the
true error. The conditional second moment of the true error can be found
from results in \cite{Zollanvari}, which for completeness are represented
here: 
\begin{equation}
\begin{aligned}
E_{S_{n}}[\varepsilon _{0}^{2}|\boldsymbol{\mu }]\eqsim \Phi \left( \frac{%
-G_{0}^{R}+c}{\sqrt{D_{0}^{R}}};\frac{C_{0}^{T,R}}{D_{0}^{R}}\right) ,
\end{aligned}
\label{eq-2dam112trssoTcon}
\end{equation}%
with $G_{0}^{R}$ and $D_{0}^{R}$ defined in (\ref{bjsbkjas}), 
\begin{equation}
\begin{aligned}
E_{S_{n}}[\varepsilon _{1}^{2}|\boldsymbol{\mu }]\eqsim \Phi \left( \frac{%
G_{1}^{R}-c}{\sqrt{D_{1}^{R}}};\frac{C_{1}^{T,R}}{D_{1}^{R}}\right) ,
\end{aligned}
\label{eq-2dam112trssoTconm}
\end{equation}%
and 
\begin{equation}
\begin{aligned}
E_{S_{n}}[\varepsilon _{0}\varepsilon _{1}|\boldsymbol{\mu }]\eqsim \Phi
\left( \frac{-G_{0}^{R}+c}{\sqrt{D_{0}^{R}}},\frac{G_{1}^{R}-c}{\sqrt{%
D_{1}^{R}}};\frac{C_{01}^{T,R}}{\sqrt{D_{0}^{R}D_{1}^{R}}}\right) ,
\end{aligned}
\label{eq-2dam112trssoTcona}
\end{equation}%
where 
\begin{equation}
C_{01}^{T,R}=-\frac{p}{2n_{0}^{2}}-\frac{p}{2n_{1}^{2}}.
\end{equation}%
Similar to obtaining (\ref{eq-2dam112trss}), we can show that 
\begin{equation}
\begin{aligned}
E_{\boldsymbol{\mu },S_{n}}[\varepsilon _{0}^{2}]\eqsim \Phi \left( \frac{%
-H_{0}^{R}+c}{\sqrt{F_{0}^{R}}};\frac{K_{0}^{T,R}}{F_{0}^{R}}\right) ,
\end{aligned}
\label{eq-2dam112trssoT}
\end{equation}%
with $H_{0}^{R}$ and $F_{0}^{R}$ given in (\ref{jacdbdx}) and (\ref{f0f}),
respectively, and 
\begin{equation}
\begin{aligned}
K_{0}^{T,R}=\left( \frac{1}{\nu _{0}}+\frac{1}{\nu _{1}}+\frac{1}{n_{1}}%
\right) \Delta _{\mathbf{m}}^{2}+\frac{p}{2\nu _{0}^{2}}+\frac{p}{2\nu
_{1}^{2}}+\frac{p}{\nu _{0}\nu _{1}}+\frac{p}{2n_{0}^{2}}+\frac{p}{2n_{1}^{2}%
}+\frac{p}{n_{1}\nu _{0}}+\frac{p}{n_{1}\nu _{1}}.
\end{aligned}
\label{k01TTf}
\end{equation}%
Similarly, 
\begin{equation}
\begin{aligned}
E_{\boldsymbol{\mu },S_{n}}[\varepsilon _{1}^{2}]\,=\,\Phi \left( \frac{%
H_{1}^{R}-c}{\sqrt{F_{1}^{R}}};\frac{K_{1}^{T,R}}{F_{1}^{R}}\right) ,
\end{aligned}
\label{eq-2dam1123sshhT}
\end{equation}%
with $K_{1}^{T,R}$ obtained from $K_{0}^{T,R}$ by exchanging $n_{0}$ and $%
n_{1}$, and $\nu _{0}$ and $\nu _{1}$. Similarly, 
\begin{equation}
\begin{aligned}
E_{\boldsymbol{\mu },S_{n}}[\varepsilon _{0}^{2}]\eqsim \Phi \left( \frac{%
-H_{0}^{R}+c}{\sqrt{F_{0}^{R}}},\frac{H_{1}^{R}-c}{\sqrt{F_{1}^{R}}};\frac{%
K_{01}^{T,R}}{\sqrt{F_{0}^{R}F_{1}^{R}}}\right) ,
\end{aligned}
\label{eq-2dam112trssoTT}
\end{equation}%
with $H_{0}^{R}$ and $F_{0}^{R}$ given in (\ref{jacdbdx}) and (\ref{f0f}),
respectively, and 
\begin{equation}
\begin{aligned}
K_{01}^{T,R}=\left( \frac{1}{\nu _{0}}+\frac{1}{\nu _{1}}\right) \Delta _{%
\mathbf{m}}^{2}+\frac{p}{2\nu _{0}^{2}}+\frac{p}{2\nu _{1}^{2}}+\frac{p}{\nu
_{0}\nu _{1}}-\frac{p}{2n_{0}^{2}}-\frac{p}{2n_{1}^{2}}.
\end{aligned}
\label{k01TTfT}
\end{equation}

\section{Monte Carlo Comparisons}

In this section we compare the asymptotically exact finite-sample
approximations of the first, second and mixed moments to Monte Carlo
estimations in conditional and unconditional scenarios. The following steps
are used to compute the Monte Carlo estimation:

\begin{itemize}
\item 
\begin{enumerate}
\item Define a set of hyper-parameters for the Gaussian model: $\mathbf{m}%
_{0} $, $\mathbf{m}_{1}$, $\nu _{0}$, ,$\nu _{1}$, and $\mathbf{\Sigma }$.
We let $\mathbf{\Sigma }$ have diagonal elements 1 and off-diagonal elements
0.1. $\mathbf{m}_{0}$ and $\mathbf{m}_{1}$ are chosen by fixing $\delta _{%
\boldsymbol{\mu }}^{2}$ ($\delta _{\boldsymbol{\mu }}^{2}=4$, which
corresponds to Bayes error $0.1586$). Setting $\delta _{\boldsymbol{\mu }%
}^{2}$ and $\mathbf{\Sigma }$ fixes the means $\boldsymbol{\mu }_{0}$ and $%
\boldsymbol{\mu }_{1}$ of the class-conditional densities (we assumed $\boldsymbol{\mu }_{i}$ has equal elements and $\boldsymbol{\mu }_{0}=-\boldsymbol{\mu }_{1}$). The priors, $\pi
_{0}$ and $\pi _{1}$, are defined by choosing a small deviation from $%
\boldsymbol{\mu }_{0}$ and $\boldsymbol{\mu }_{1}$, that is, by setting $%
\mathbf{m}_{i}=\boldsymbol{\mu }_{i}+a\boldsymbol{\mu }_{i}$, where $a=0.01$.

\item (unconditional case): Using $\pi _{0}$ and $\pi _{1}$, generate random
realizations of $\mathbf{\boldsymbol{\mu }}_{0}$ and $\mathbf{\boldsymbol{%
\mu }}_{1}$.

\item (conditional case): Use the values of $\boldsymbol{\mu }_{0}$ and $%
\boldsymbol{\mu }_{1}$ obtained from Step 1.

\item For fixed $\Pi _{0}$ and $\Pi _{1}$, generate a set of training data
of size $n_{i}$ for class $i=0,1$.

\item Using the training sample, design the LDA classifier, $\psi _{n}$,
using (\ref{LDAc}).

\item Compute the Bayesian MMSE error estimator, $\hat{\varepsilon}^{B}$,
using (\ref{eq:BEEU}) and (\ref{qwsaqwsaU}).

\item Knowing $\mathbf{\boldsymbol{\mu }}_{0}$ and $\mathbf{\boldsymbol{\mu }%
}_{1}$, find the true error of $\psi _{n}$ using (\ref{eq:true_errorU}).

\item Repeat Steps 3 through 6, $T_{1}$ times.

\item Repeat Steps 2 through 7, $T_{2}$ times.
\end{enumerate}
\end{itemize}

In the unconditional case, we set $T_{1}=T_{2}=300$ and generate $90,000$
samples. For the conditional case, we set $T_{1}=10,000$ and $T_{2}=1$, the
latter because $\mathbf{\boldsymbol{\mu }}_{0}$ and $\mathbf{\boldsymbol{\mu 
}}_{1}$ are set in Step 2.

Figure 1 treats Raudys-type
finite-sample approximations, including the RMS. Figure 1(a)\ compares the
first moments obtained from equations (\ref{condRaudys}) and (\ref%
{uncondRaudys}). It presents $E_{S_{n}}[{\hat{\varepsilon}}^{B}|\boldsymbol{%
\mu }]$ and $E_{\boldsymbol{\mu },S_{n}}[{\hat{\varepsilon}}^{B}]$ computed
by Monte Carlo estimation and the analytical expressions. The label
\textquotedblleft FSA BE Uncond" identifies the curve of $E_{\boldsymbol{\mu 
},S_{n}}[{\hat{\varepsilon}}^{B}]$, the unconditional expected estimated
error obtained from the finite-sample approximation, which according to the
basic theory is equal to $E_{\boldsymbol{\mu },S_{n}}[{\varepsilon }]$. The
labels \textquotedblleft FSA BE Cond" and \textquotedblleft FSA TE Cond"
show the curves of $E_{S_{n}}[{\hat{\varepsilon}}^{B}|\boldsymbol{\mu }]$,
the conditional expected estimated error, and $E_{S_{n}}[{\varepsilon }|%
\boldsymbol{\mu }]$, the conditional expected true error, respectively, both
obtained using the analytic approximations. The curves obtained from Monte
Carlo estimation are identified by \textquotedblleft MC" labels. The
analytic curves in Figure 1(a) show substantial agreement with the Monte
Carlo approximation.

To obtain the second moments, $\text{Var}^{d}[\hat{\varepsilon}]$ and $\text{%
RMS}[\hat{\varepsilon}^{B}]$ as defined in (\ref{eq-RMS}), we use equations (%
\ref{eq-2dam1123ss}), (\ref{eq-2dam1123s}), (\ref{eq-2dam112u}), (\ref%
{eq-2dam112gb}), (\ref{eq-2dam112gcb}), (\ref{eq-2dam112trssoTcon}), (\ref%
{eq-2dam112trssoTconm}), (\ref{eq-2dam112trssoTcona}) for the conditional
case and (\ref{eq-2dam1123sskk}), (\ref{eq-2dam1123sshh}), (\ref%
{eq-2dam112oo}), (\ref{eq-2dam112trss}), (\ref{eq-2dam112trsso}), (\ref%
{eq-2dam112trssoT}), (\ref{eq-2dam1123sshhT}), (\ref{eq-2dam112trssoTT}) for
the unconditional case. Figures 1(b), 1(c), and 1(d) compare the Monte Carlo
estimation to the finite-sample approximations obtained for second/mixed
moments, $\text{Var}^{d}[\hat{\varepsilon}]$, and $\text{RMS}[\hat{%
\varepsilon}^{B}]$, respectively. The labels are interpreted similarly to
those in Figure 1(a), but for the second/mixed moments instead. For example,
\textquotedblleft MC BE$\times $TE Uncond" identifies the MC curve of $E_{%
\boldsymbol{\mu },S_{n}}[\hat{\varepsilon}^{B}\varepsilon ]$. The Figures
1(b), 1(c), and 1(d) show that the finite-sample approximations for the
conditional and unconditional second/mixed moments, variance of deviation,
and RMS are quite accurate (close to the MC value).

While Figure 1 shows the accuracy of Raudys-type of finite-sample
approximations, figures in the Supplementary Materials show the the
comparison between the finite-sample approximations obtained directly from
Theorem 1-6, i.e. equations (29), (57), (70), (73), (76), (102), and (103),
to Monte Carlo estimation.

\section{Examination of the Raudys-type RMS Approximation}

Equations (\ref{g0f}), (\ref{d0f}), (\ref{c0f}), (\ref{c01f}), and (\ref%
{eq-2dam112gcb}) show that $\text{RMS}_{S_{n}}[\hat{\varepsilon}^{B}|%
\boldsymbol{\mu }]$ is a function of 14 variables: \newline
$p,n_{0},n_{1},\beta _{0},\beta _{1},\delta _{\boldsymbol{\mu }}^{2},\eta _{%
\mathbf{m}_{0},\boldsymbol{\mu }_{1}},\eta _{\mathbf{m}_{0},\boldsymbol{\mu }%
_{0}},\eta _{\mathbf{m}_{1},\boldsymbol{\mu }_{0}},\eta _{\mathbf{m}_{1},%
\boldsymbol{\mu }_{1}},\eta _{\mathbf{m}_{0},\boldsymbol{\mu }_{0},%
\boldsymbol{\mu }_{0},\boldsymbol{\mu }_{1}},\eta _{\mathbf{m}_{0},%
\boldsymbol{\mu }_{0},\mathbf{m}_{1},\boldsymbol{\mu }_{0}},\eta _{\mathbf{m}%
_{1},\boldsymbol{\mu }_{1},\mathbf{m}_{0},\boldsymbol{\mu }_{1}},\eta _{%
\mathbf{m}_{1},\boldsymbol{\mu }_{1},\boldsymbol{\mu }_{1},\boldsymbol{\mu }%
_{0}}$. Studying a function of this number of variables is complicated,
especially because restricting some variables can constrain others. We make
several simplifying assumptions to reduce the complexity. We let $%
n_{0}=n_{1}=\frac{n}{2}$, $\beta _{0}=\beta _{1}=\beta $ and assumed priors are centered at unknown true means, i.e. $\mathbf{m}_{0}=\boldsymbol{\mu }_{0}$ and $%
\mathbf{m}_{1}=\boldsymbol{\mu }_{1}$. Using these assumptions, $\text{RMS}%
_{S_{n}}[\hat{\varepsilon}^{B}|\boldsymbol{\mu }]$ is only a function of $%
p,n,\beta $, and $\delta _{\boldsymbol{\mu }}^{2}$. We let $p\in \lbrack
4,200]$, $n\in \lbrack 40,200]$, $\beta \in \{0.5,1,2\}$, $\delta _{%
\boldsymbol{\mu }}^{2}\in \{4,16\}$, which means that the Bayes error is $%
0.158$ or $0.022$. Figure 2(a) shows plots of $\text{RMS}_{S_{n}}[\hat{%
\varepsilon}^{B}|\boldsymbol{\mu }]$ as a function of $p$, $n$, $\beta $,
and $\delta _{\boldsymbol{\mu }}^{2}$. These show that for smaller distance
between classes, that is, for smaller $\delta _{\boldsymbol{\mu }}^{2}$
(larger Bayes error), the R$\text{MS}$ is larger, and as the distance
between classes increases, the $\text{RMS}$ decreases. Furthermore, we see
that in situations where very informative priors are available, i.e. $%
\mathbf{m}_{0}=\boldsymbol{\mu }_{0}$ and $\mathbf{m}_{1}=\boldsymbol{\mu }%
_{1}$, relying more on data can have a detrimental effect on $\text{RMS}$.
Indeed, the plots in the top row (for $\beta =0.5$) have larger $\text{RMS}$
than the plots in the bottom row of the figure (for $\beta =2$).

Using the RMS expressions enables finding the necessary sample size to
insure a given $\text{RMS}_{S_{n}}[\hat{\varepsilon}^{B}|\boldsymbol{\mu }]$
by using the same methodology as developed for the resubstitution and
leave-one-out error estimators in \cite{Zollanvari,ILLUSION}. The plots in
Figure 2(a) (as well as other unshown plots) show that, with $\mathbf{m}_{0}=%
\boldsymbol{\mu }_{0}$ and $\mathbf{m}_{1}=\boldsymbol{\mu }_{1}$, the $%
\text{RMS}$ is a decreasing function of $\delta _{\boldsymbol{\mu }}^{2}$.
Therefore, the number of sample points that guarantees $\max_{\delta _{%
\boldsymbol{\mu }}^{2}>0}\text{RMS}_{S_{n}}[\hat{\varepsilon}^{B}|%
\boldsymbol{\mu }]=\lim_{\delta _{\boldsymbol{\mu }}^{2}\rightarrow 0}\text{%
RMS}_{S_{n}}[\hat{\varepsilon}^{B}|\boldsymbol{\mu }]$ being less than a
predetermined value $\tau $ insures that $\text{RMS}_{S_{n}}[\hat{\varepsilon%
}^{B}|\boldsymbol{\mu }]<\tau ,$ for any $\delta _{\boldsymbol{\mu }}^{2}$.
Let the desired bound be $\kappa _{\hat{\varepsilon}}(n,p,\beta
)=\lim_{\delta _{\boldsymbol{\mu }}^{2}\rightarrow 0}\text{RMS}_{S_{n}}[\hat{%
\varepsilon}^{B}|\boldsymbol{\mu }]$. From equations (\ref{eq-2dam1123ss}), (%
\ref{eq-2dam1123s}), (\ref{eq-2dam112u}), (\ref{eq-2dam112gb}), (\ref%
{eq-2dam112gcb}), (\ref{eq-2dam112trssoTcon}), (\ref{eq-2dam112trssoTconm}),
and (\ref{eq-2dam112trssoTcona}), we can find $\kappa _{\hat{\varepsilon}%
}(n,p,\beta )$ and increase $n$ until $\kappa _{\hat{\varepsilon}}(n,p,\beta
)<\tau $. Table 1 ($\beta =1$: Conditional) shows the minimum number of
sample points needed to guarantee having a predetermined conditional $\text{%
RMS}$ for the whole range of $\delta _{\boldsymbol{\mu }}^{2}$ (other $\beta 
$ shown in the Supplementary Material). A larger dimensionality, a smaller $%
\tau $, and a smaller $\beta $ result in a larger necessary sample size
needed for having $\kappa _{\hat{\varepsilon}}(n,p,\beta )<\tau $.

Turning to the unconditional RMS, equations (\ref{f0f}), (\ref{k0f}), (\ref%
{k01f}), (\ref{k01Tf}), (\ref{k01TTf}), and (\ref{k01TTfT}) show that $\text{%
RMS}_{\boldsymbol{\mu },S_{n}}[\hat{\varepsilon}^{B}]$ is a function of 6
variables: $p,n_{0},n_{1},\nu _{0},\nu _{1},\Delta _{\mathbf{m}}^{2}$.
Figure 2(b) shows plots of $\text{RMS}_{\boldsymbol{\mu },S_{n}}[\hat{%
\varepsilon}^{B}]$ as a function of $p$, $n$, $\beta $, and $\Delta _{%
\mathbf{m}}^{2}$, assuming $n_{0}=n_{1}=\frac{n}{2}$, $\beta _{0}=\beta
_{1}=\beta $. Note that setting the values of $n$ and $\beta $ fixes the
value of $\nu _{0}=\nu _{1}=\nu $ in the corresponding expressions for $%
\text{RMS}_{\boldsymbol{\mu },S_{n}}[\hat{\varepsilon}^{B}]$. Due to the
complex shape of $\text{RMS}_{\boldsymbol{\mu },S_{n}}[\hat{\varepsilon}%
^{B}] $, we consider a large range of $n$ and $p$. The plots show that a
smaller distance between prior distributions (smaller $\Delta _{\mathbf{m}%
}^{2}$) corresponds to a larger unconditional $\text{RMS}$ of estimation. In
addition, as the distance between classes increases, the $\text{RMS}$
decreases. The plots in Figure 2(b) show that, as $\Delta _{\mathbf{m}}^{2}$
increases, $\text{RMS}$ decreases. Furthermore, Figure 2(b) (and other
unshown plots) demonstrate an interesting phenomenon in the shape of the $%
\text{RMS}$. In regions defined by pairs of $(p,n)$, for each $p$, $\text{RMS%
}$ first increases as a function of sample size and then decreases. We
further observe that with fixed $p$, for smaller $\beta $, this
\textquotedblleft peaking phenomenon" happens for larger $n$. On the other
hand, with fixed $\beta $, for larger $p$, peaking happens for larger $n$.
These observations are presented in Figure 3, which shows curves obtained by
cutting the 3D plots in the left column of Fig. 2(b) at a few dimensions.
This figure shows that, for $p=900$ and $\beta =2$, adding more sample
points increases $\text{RMS}$ abruptly at first to reach a maximum value of $%
\text{RMS}$ at $n=140$, the point after which the $\text{RMS}$ starts to
decrease.

One may use the unconditional scenario to determine the the minimum
necessary sample size for a desired $\text{RMS}_{\boldsymbol{\mu },S_{n}}[%
\hat{\varepsilon}^{B}]$. In fact, this is the more practical way to go
because in practice one does not know $\boldsymbol{\mu }$. Since the
unconditional $\text{RMS}$ shows a decreasing trend in terms of $\Delta _{%
\mathbf{m}}^{2}$, we use the previous technique to find the minimum
necessary sample size to guarantee a desired unconditional $\text{RMS}$.
Table 1 ($\beta =1$: Unconditional) shows the minimum sample size that
guarantees $\max_{\Delta _{\mathbf{m}}^{2}>0}\text{RMS}_{\boldsymbol{\mu }%
,S_{n}}[\hat{\varepsilon}^{B}]=\lim_{\Delta _{\mathbf{m}}^{2}\rightarrow 0}%
\text{RMS}_{\boldsymbol{\mu },S_{n}}[\hat{\varepsilon}^{B}]$ being less than
a predetermined value $\tau $, i.e. insures that $\text{RMS}_{\boldsymbol{%
\mu },S_{n}}[\hat{\varepsilon}^{B}]<\tau $ for any $\Delta _{\mathbf{m}}^{2}$
(other $\beta $ shown in the Supplementary Material).

To examine the accuracy of the required sample size that satisfies $\kappa _{%
\hat{\varepsilon}}(n,p,\beta )<\tau $ for both conditional and unconditional
settings, we have performed a set of experiments (see Supplementary
Material). The results of these experiments confirm the efficacy of Table 1
in determining the minimum sample size required to insure the RMS is less
than a predetermined value $\tau $.

\begin{figure}[htp]
\centering
\addtolength{\subfigcapskip}{-0.2in}%
\subfigure[]{
\includegraphics[scale=0.77]{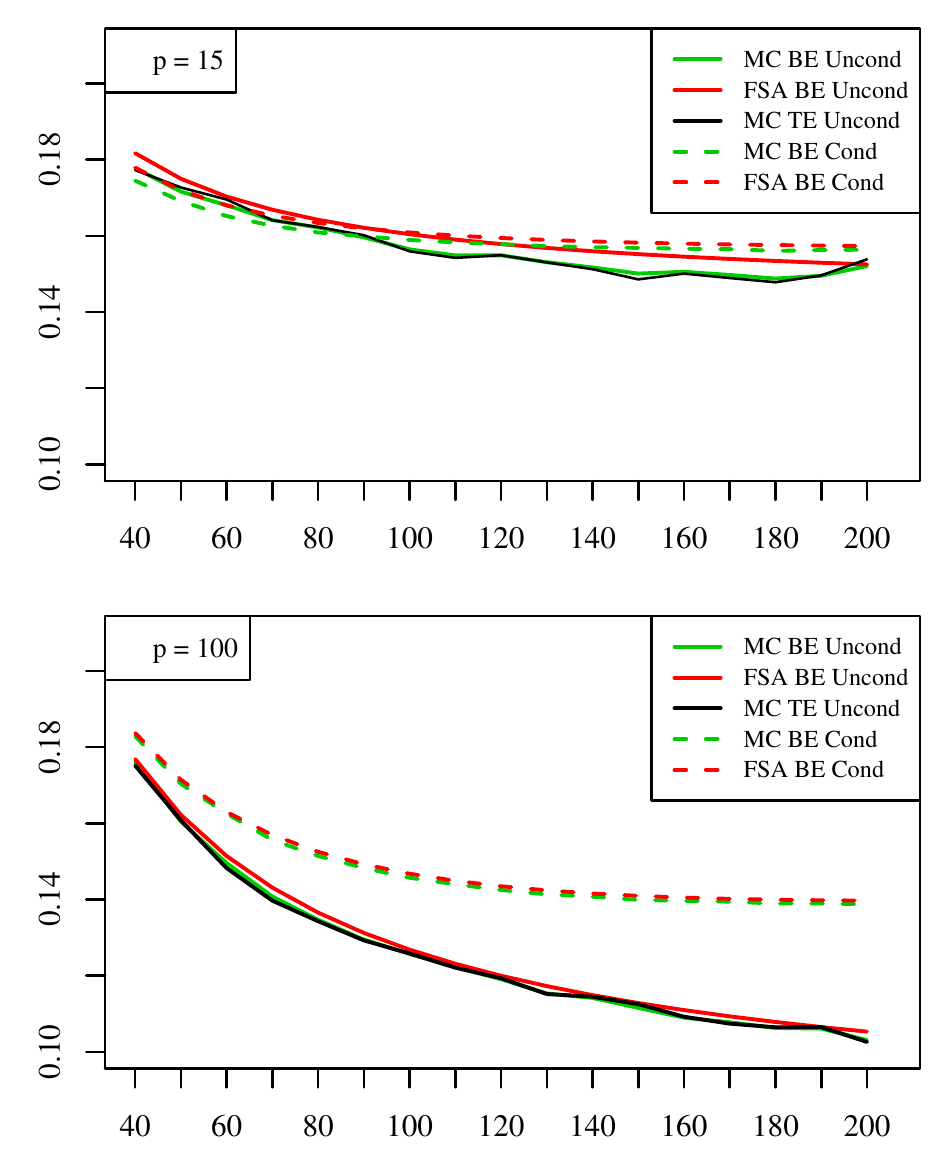}
\label{Fig-ex2a}
}\hspace{0.5cm} 
\subfigure[]{
\includegraphics[scale=0.77]{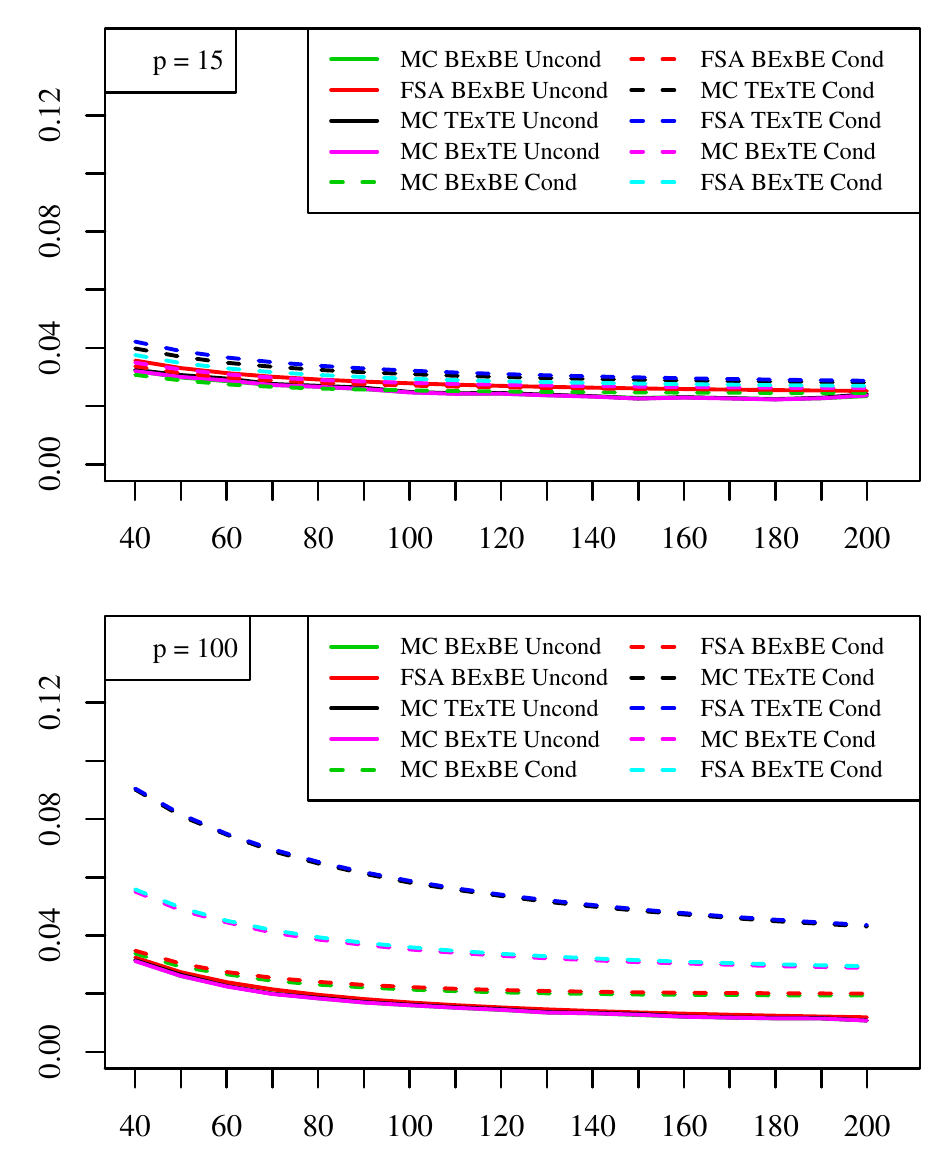}
\label{Fig-ex2b}
} 
\subfigure[]{
\includegraphics[scale=0.77]{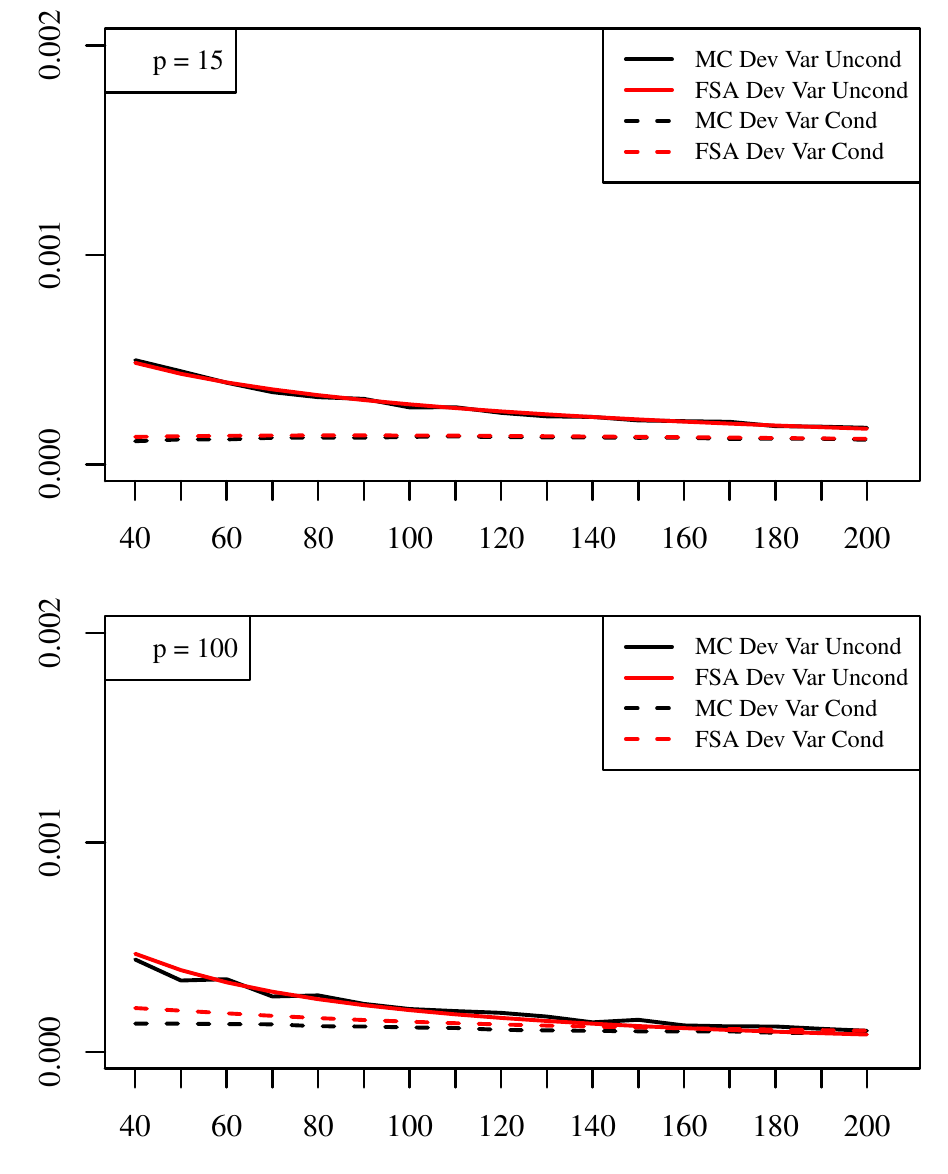}
\label{Fig-ex2c}
}\hspace{0.5cm} 
\subfigure[]{
\includegraphics[scale=0.77]{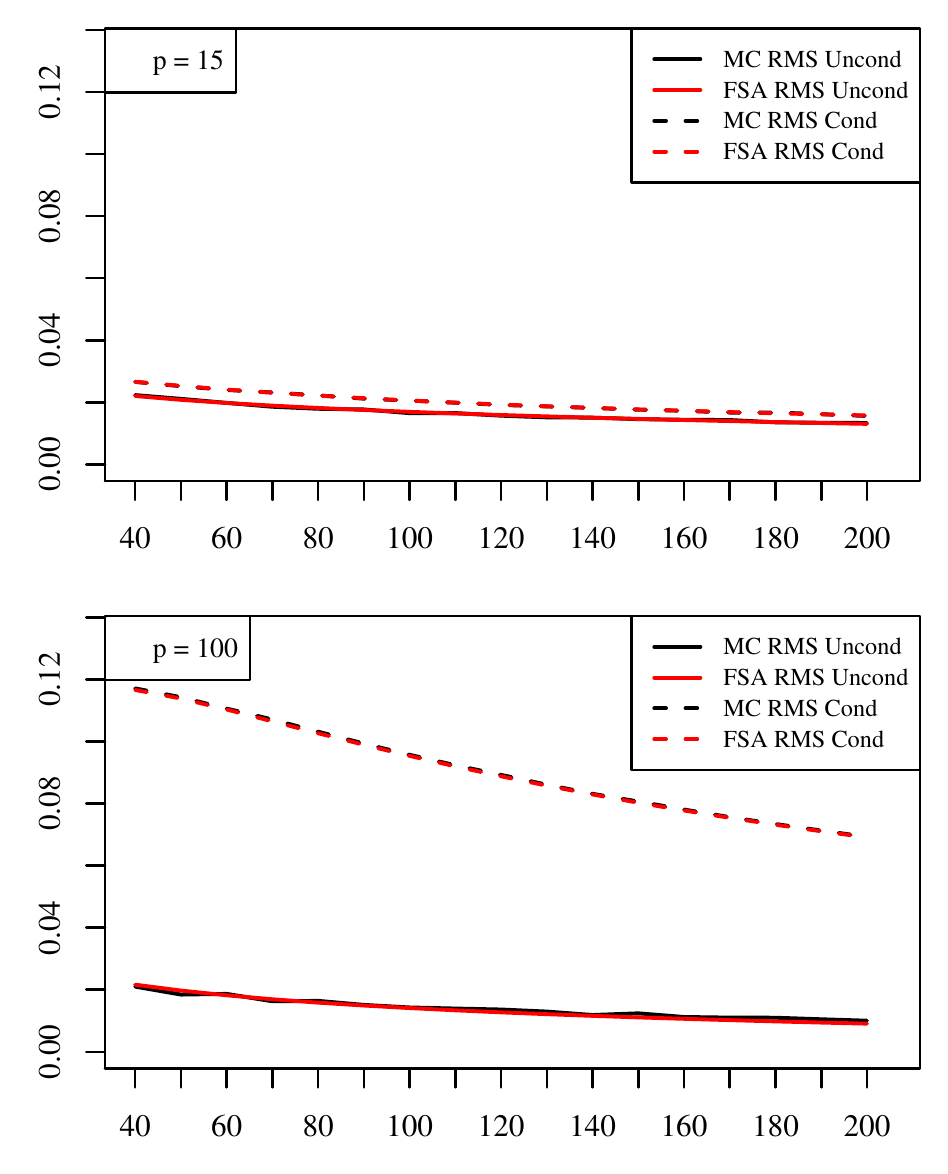}
\label{Fig-ex2d}
}\vspace{-0.3cm}
\caption{Comparison of conditional and unconditional performance metrics of $%
{\hat{\protect\varepsilon}}^B$ using asymptotically exact finite setting
approximations, with Monte Carlo estimates as a function of sample size. (a)
Expectations. The case of asymptotic unconditional expectation of ${\protect%
\varepsilon}$ is not plotted as ${\hat{\protect\varepsilon}}^B$ is
unconditionally unbiased; (b) Second and mixed moments; (c) Conditional
variance of deviation from true error, i.e. $\text{Var}^{d}_{S_n}[\hat{%
\protect\varepsilon}^B|\boldsymbol{\protect\mu}]$ and, unconditional
variance of deviation, i.e. $\text{Var}^{d}_{\boldsymbol{\protect\mu},S_n}[%
\hat{\protect\varepsilon}^B]$; (d) Conditional $\text{RMS}$ of estimation,
i.e. $\text{RMS}_{S_n}[\hat{\protect\varepsilon}^B|\boldsymbol{\protect\mu}]$
and, unconditional $\text{RMS}$ of estimation, i.e. $\text{RMS}_{\boldsymbol{%
\protect\mu},S_n}[\hat{\protect\varepsilon}^B]$; (a)-(d) correspond to the
same scenario in which dimension, $p$, is 15 and 100, $\protect\nu_0=\protect%
\nu_1=50$, $\mathbf{m}_i=\boldsymbol{\protect\mu}_i+0.01\boldsymbol{\protect%
\mu}_i$ with $\boldsymbol{\protect\mu}_0=-\boldsymbol{\protect\mu}_1$, and
Bayes error = 0.1586. }
\label{Fig-ex2}
\end{figure}

%
\begin{figure}[ht]
\centering
\subfigure[]{
\includegraphics[scale=0.800]{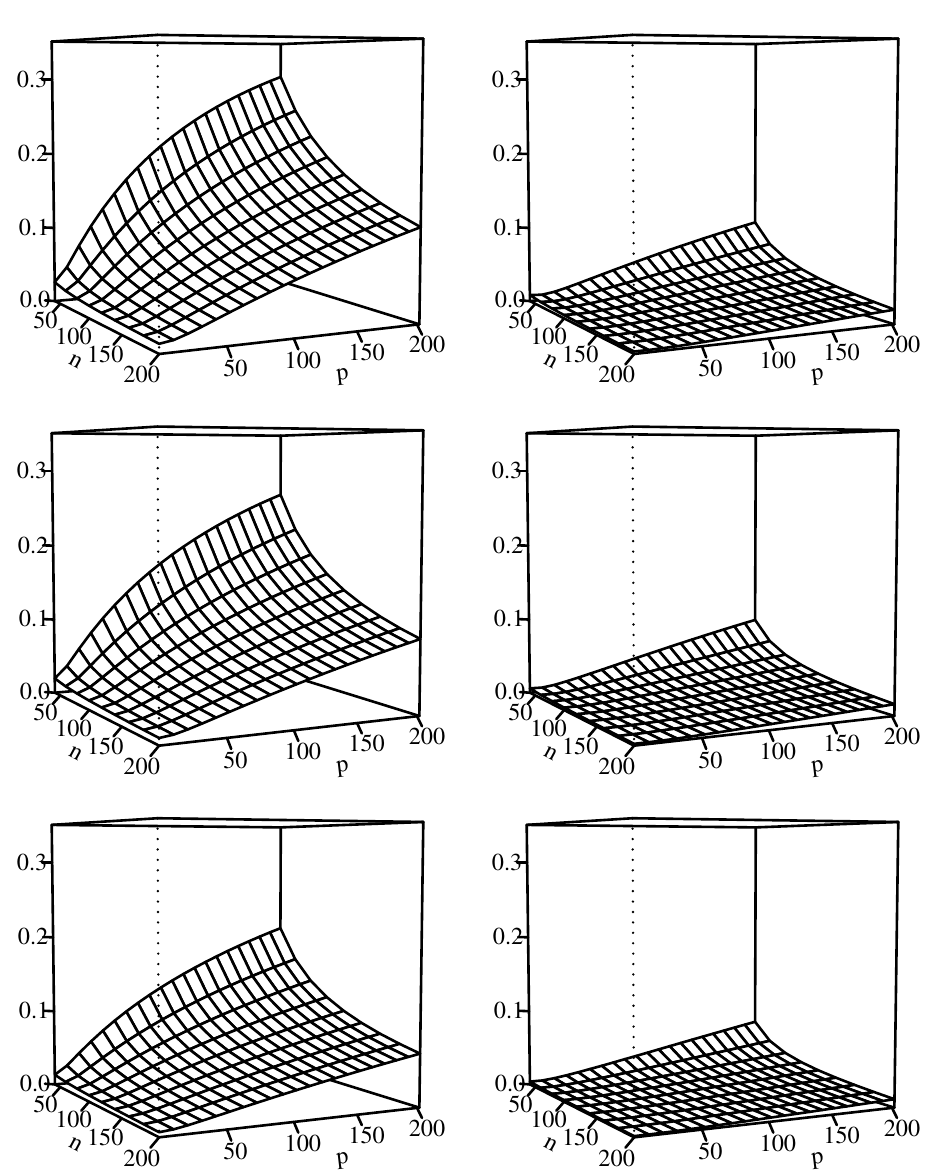}
\label{Fig-ex5a}
}\hspace{0.45cm} 
\subfigure[]{
\includegraphics[scale=0.800]{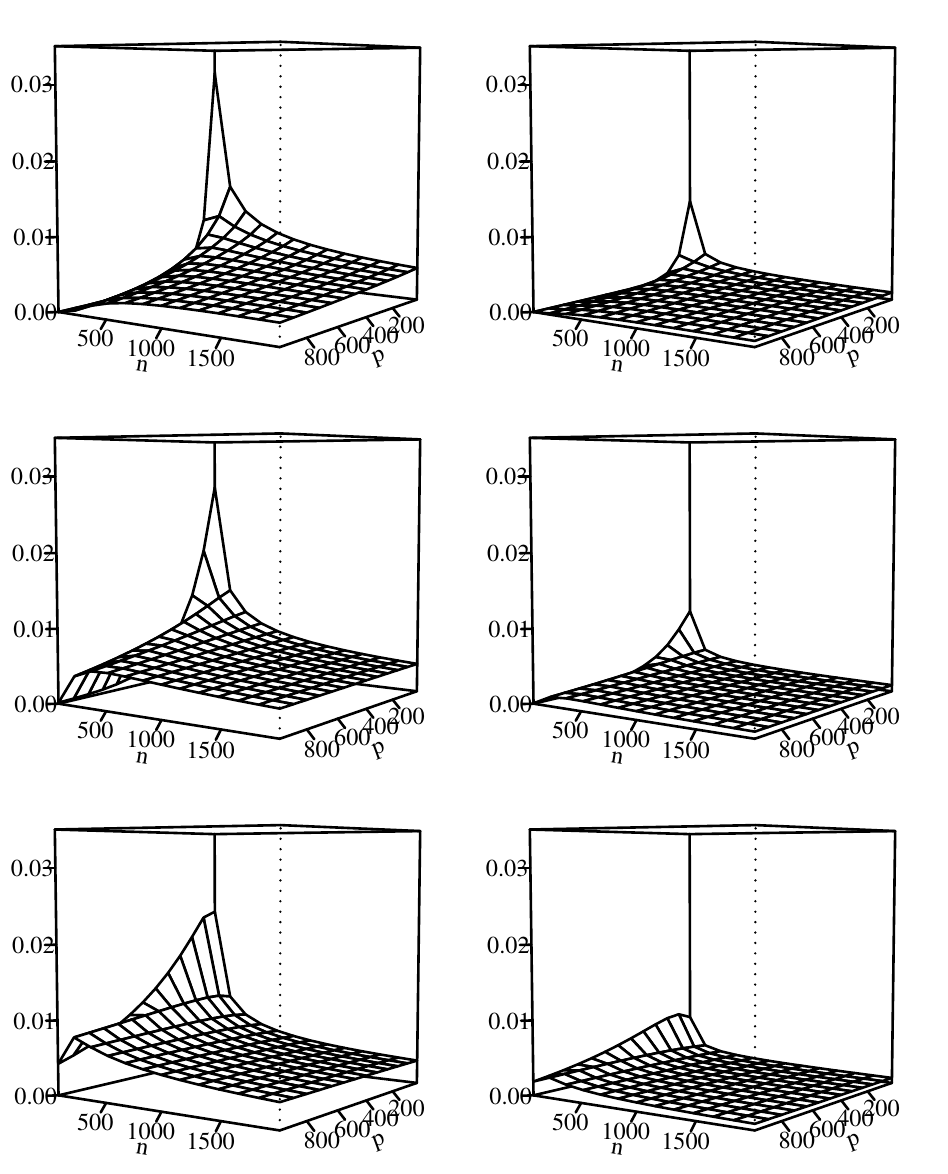}
\label{Fig-ex5b}
}
\caption{(a) The conditional $\text{RMS}$ of estimation, i.e. $\text{RMS}%
_{S_n}[\hat{\protect\varepsilon}^B|\boldsymbol{\protect\mu}]$, as a function
of $p<200$ and $n<200$. From top to bottom, the rows correspond to $\protect%
\beta=0.5, 1, 2$, respectively. From left to right, the columns correspond
to $\protect\delta^2_{\boldsymbol{\protect\mu}}=4, 16$, respectively. (b)
The unconditional $\text{RMS}$ of estimation, i.e. $\text{RMS}_{\boldsymbol{%
\protect\mu},S_n}[\hat{\protect\varepsilon}^B]$, as a function of $p<1000$
and $n<2000$. From top to bottom, the rows correspond to $\protect\beta=0.5,
1, 2$, respectively. From left to right, the columns correspond to $\Delta^2_%
\mathbf{m}=4, 16$, respectively.}
\label{Fig-ex5}
\end{figure}

\begin{figure}[t]
\centering
\includegraphics[scale=0.95]{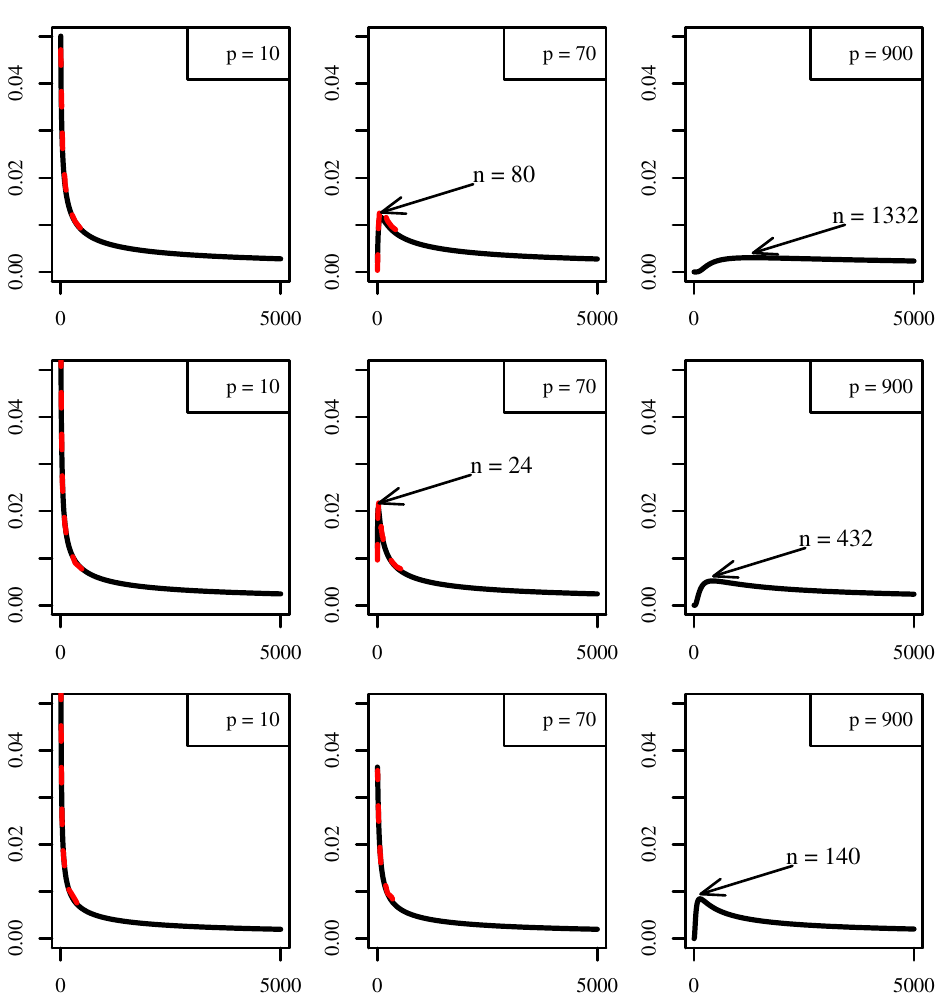}
\caption{$\text{RMS}_{\boldsymbol{\protect\mu},S_n}[\hat{\protect\varepsilon}%
^B]$-peaking phenomenon as a function of sample size. These plots are
obtained by cutting the 3D plots in the left column of Fig. 2(b) at few
dimensionality (i.e. $\Delta^2_\mathbf{m}=4$). From top to bottom the rows
correspond to $\protect\beta=0.5, 1, 2$, respectively. The solid-black
curves indicate $\text{RMS}_{\boldsymbol{\protect\mu},S_n}[\hat{\protect%
\varepsilon}^B]$ computed from the analytical results and the red-dashed
curves show the same results computed by means of Monte Carlo simulations.
Due to computational burden of estimating the curves by means of Monte Carlo
studies, the simulations are limited to $n<500$ and $p=10,70$. }
\label{Fig-ex7}
\end{figure}
\begin{table}[!tbp]
\caption{Minimum sample size, $n$, ($n_0=n_1=\frac{n}{2}$) to satisfy $\protect\kappa_{\hat{\protect\varepsilon}}(n, p,\beta)<\tau$.  }
\tiny
\begin{center}
\begin{tabular}{rrrrrrrr}
\hline\hline
\multicolumn{1}{r}{$\tau$}&\multicolumn{1}{c}{p = 2}&\multicolumn{1}{c}{p = 4}&\multicolumn{1}{c}{p = 8}&\multicolumn{1}{c}{p = 16}&\multicolumn{1}{c}{p = 32}&\multicolumn{1}{c}{p = 64}&\multicolumn{1}{c}{p = 128}\tabularnewline
\hline
{\bfseries $\beta=1$: Conditional}&&&&&&&\tabularnewline
0.1&$ 14$&$ 22$&$ 38$&$ 70$&$132$&$ 256$&$ 506$\tabularnewline
0.09&$ 18$&$ 28$&$ 48$&$ 86$&$164$&$ 318$&$ 626$\tabularnewline
0.08&$ 24$&$ 36$&$ 60$&$110$&$208$&$ 404$&$ 796$\tabularnewline
0.07&$ 32$&$ 48$&$ 80$&$144$&$272$&$ 530$&$1044$\tabularnewline
0.06&$ 44$&$ 64$&$108$&$196$&$372$&$ 722$&$1424$\tabularnewline
0.05&$ 62$&$ 94$&$158$&$284$&$538$&$1044$&$2056$\tabularnewline
\hline
{\bfseries $\beta=1$: Unconditional}&&&&&&&\tabularnewline
0.025&$ 108$&$ 108$&$ 106$&$ 102$&$  92$&$  72$&$   2$\tabularnewline
0.02&$ 172$&$ 170$&$ 168$&$ 164$&$ 156$&$ 138$&$  78$\tabularnewline
0.015&$ 308$&$ 306$&$ 304$&$ 300$&$ 292$&$ 274$&$ 236$\tabularnewline
0.01&$ 694$&$ 694$&$ 690$&$ 686$&$ 678$&$ 662$&$ 628$\tabularnewline
0.005&$2790$&$2786$&$2782$&$2776$&$2768$&$2752$&$2720$\tabularnewline
\hline
\end{tabular}
\end{center}
\label{table:ex1}
\end{table}

\section{Conclusion}

Using realistic assumptions about sample size and dimensionality, standard
statistical techniques are generally incapable of estimating the error of a
classifier in small-sample classification. Bayesian MMSE error estimation
facilitates more accurate estimation by incorporating prior knowledge. In
this paper, we have characterized two sets of performance metrics for
Bayesian MMSE error estimation in the case of LDA in a Gaussian model: (1)
the first, second, and cross moments of the estimated and actual errors
conditioned on a fixed feature-label distribution, which in turn gives us
knowledge of the conditional RMS$_{S_{n}}[\hat{\varepsilon}^{B}|\theta ]$;
and (2) the unconditional moments and, therefore, the unconditional RMS, RMS$%
_{\theta ,S_{n}}[\hat{\varepsilon}^{B}]$. We set up a series of conditions,
called the Bayesian-Kolmogorov asymptotic conditions, that allow us to
characterize the performance metrics of Bayesian MMSE error estimation in an
asymptotic sense. The Bayesian-Kolmogorov asymptotic conditions are set up
based on the assumption of increasing $n$, $p$, and certainty parameter $\nu 
$, with an arbitrary constant limiting ratio between $n$ and $p$, and $n$
and $\nu $. To our knowledge, these conditions permit, for the first time,
application of Kolmogorov-type of asymptotics in a Bayesian setting. The
asymptotic expressions proposed in this paper result directly in
finite-sample approximations of the performance metrics. Improved
finite-sample accuracy is achieved via newly proposed Raudys-type
approximations. The asymptotic theory is used to prove that these
approximations are, in fact, asymptotically exact under the
Bayesian-Kolmogorov asymptotic conditions. \noindent Using the derived
analytical expressions, we have examined performance of the Bayesian MMSE
error estimator in relation to feature-label distributions, prior knowledge,
sample size, and dimensionality. We have used the results to determine the
minimum sample size guaranteeing a desired level of error estimation
accuracy.

As noted in the Introduction, a natural next step in error estimation theory
is to remove the known-covariance condition, but as also noted, this may
prove to be difficult.
\vspace{-0.2cm}
\section{Acknowledgments}

This work was partially supported by the NIH grants 2R25CA090301 (Nutrition,
Biostatistics, and Bioinformatics) from the National Cancer Institute. We would like to thank Dr. Lori Dalton for her critical review of the manuscript. 

\vspace{-0.2cm}
\section*{Appendix}

\section*{Proof of Theorem 1}

We explain this proof in detail as some steps will be used in later proofs.
Let 
\begin{equation}
\hat{G}_{0}^{B}=\left( \mathbf{m}_{0}^{\ast }-\frac{\bar{\mathbf{x}}_{0}+%
\bar{\mathbf{x}}_{1}}{2}\right) ^{T}\mathbf{\Sigma }^{-1}\left( \bar{\mathbf{%
x}}_{0}-\bar{\mathbf{x}}_{1}\right) \,,  \label{pppl}
\end{equation}%
where $\mathbf{m}_{0}^{\ast }$ is defined in (\ref{mnuU}). Then 
\begin{equation}
\begin{aligned}
\!\!\hat{G}_{0}^{B}\!=\!\frac{\nu _{0}\mathbf{m}_{0}^{T}}{n_{0}+\nu _{0}}\mathbf{%
\Sigma }^{-1}(\bar{\mathbf{x}}_{0}\!-\!\bar{\mathbf{x}}_{1})+\frac{n_{0}-\nu _{0}%
}{2(n_{0}+\nu _{0})}\left( \bar{\mathbf{x}}_{0}^{T}\mathbf{\Sigma }^{-1}\bar{%
\mathbf{x}}_{0}-\bar{\mathbf{x}}_{0}^{T}\mathbf{\Sigma }^{-1}\bar{\mathbf{x}}%
_{1}\right) \!+\!\frac{1}{2}\left( \bar{\mathbf{x}}_{1}^{T}\mathbf{\Sigma }^{-1}%
\bar{\mathbf{x}}_{1}\!-\!\bar{\mathbf{x}}_{0}^{T}\mathbf{\Sigma }^{-1}\bar{%
\mathbf{x}}_{1}\right) .
\end{aligned}
\label{ghat0}
\end{equation}%
For $i,j=0,1$ and $i\neq j$, define the following random variables:

\begin{equation}
\begin{aligned}
y_{i}=\mathbf{m}_{i}^{T}\mathbf{\Sigma }^{-1}(\bar{\mathbf{x}}_{0}-\bar{%
\mathbf{x}}_{1}),\mathrm{\ \ \ }z_{i}=\bar{\mathbf{x}}_{i}^{T}\mathbf{\Sigma 
}^{-1}\bar{\mathbf{x}}_{i},\mathrm{\ \ \ }\;z_{ij}=\bar{\mathbf{x}}_{i}^{T}%
\mathbf{\Sigma }^{-1}\bar{\mathbf{x}}_{j}.
\end{aligned}
\label{ggg}
\end{equation}%
The variance of $y_{i}$ given $\boldsymbol{\mu }$ does not depend on $%
\boldsymbol{\mu }$. Therefore, under the Bayesian-Kolmogorov conditions
stated in (\ref{KACLU}), $\overline{\mathbf{m}_{i}^{T}\mathbf{\Sigma }^{-1}%
\boldsymbol{\mu }_{j}}$ and $\overline{\boldsymbol{\mu }_{i}^{T}\mathbf{%
\Sigma }^{-1}\boldsymbol{\mu }_{j}}$ do not appear in the limit. Only $%
\overline{\mathbf{m}_{i}^{T}\mathbf{\Sigma }^{-1}\mathbf{m}_{i}}$ matters,
which vanishes in the limit as follows: 
\begin{equation}
\begin{aligned}
\text{Var}_{S_{n}}[y_{i}|\boldsymbol{\mu }]=\mathbf{m}_{i}^{T}(\frac{\mathbf{%
\Sigma }^{-1}}{n_{0}}+\frac{\mathbf{\Sigma }^{-1}}{n_{1}})\mathbf{m}_{i}\,{%
\overset{K}{\rightarrow }}\;\lim_{\substack{ {n_{0}\rightarrow \infty }}}%
\frac{\overline{\mathbf{m}_{i}^{T}\mathbf{\Sigma }^{-1}\mathbf{m}_{i}}}{n_{0}%
}+\lim_{\substack{ {n_{1}\rightarrow \infty }}}\frac{\overline{\mathbf{m}%
_{i}^{T}\mathbf{\Sigma }^{-1}\mathbf{m}_{i}}}{n_{1}}=0\,.
\end{aligned}
\label{varGj}
\end{equation}

To find the variance of $z_{i}$ and $z_{ij}$ we can first transform $z_{i}$
and $z_{ij}$ to quadratic forms and then use the results of \cite{Kan:08} to
find the variance of quadratic functions of Gaussian random variables.
Specifically, from \cite{Kan:08}, for $\mathbf{y}\sim N(\boldsymbol{\mu },%
\mathbf{\Sigma })$ and $\mathbf{A}$ being a symmetric positive definite
matrix, Var$[\mathbf{y}^{T}\mathbf{A}\mathbf{y}]=2tr(\mathbf{A}\mathbf{%
\Sigma })^{2}+4\boldsymbol{\mu }^{T}\mathbf{A}\mathbf{\Sigma }\mathbf{A}%
\boldsymbol{\mu }^{\prime }$, with $tr$ being the trace operator. Therefore,
after some algebraic manipulations, we obtain 
\begin{equation}
\begin{aligned}
& \text{Var}_{S_{n}}[z_{i}|\boldsymbol{\mu }]=2\frac{p}{n_{i}^{2}}+4\frac{%
\boldsymbol{\mu }_{i}^{T}\mathbf{\Sigma }^{-1}\boldsymbol{\mu }_{i}}{n_{i}}\;%
{\overset{K}{\rightarrow }}\;2\lim_{\substack{ {n_{i}\rightarrow \infty }}}%
\frac{J_{i}}{n_{i}}+4\lim_{\substack{ {n_{i}\rightarrow \infty }}}\frac{%
\overline{\boldsymbol{\mu }_{i}^{T}\mathbf{\Sigma }^{-1}\boldsymbol{\mu }_{i}%
}}{n_{i}}=0\,, \\
& \text{Var}_{S_{n}}[z_{ij}|\boldsymbol{\mu }]=\frac{p}{n_{i}n_{j}}+\frac{%
\boldsymbol{\mu }_{i}^{T}\mathbf{\Sigma }^{-1}\boldsymbol{\mu }_{i}}{n_{j}}+%
\frac{\boldsymbol{\mu }_{j}^{T}\mathbf{\Sigma }^{-1}\boldsymbol{\mu }_{j}}{%
n_{i}}\;{\overset{K}{\rightarrow }}\;\lim_{\substack{ {n_{j}\rightarrow
\infty }}}\frac{J_{i}}{n_{j}}+\lim_{\substack{ {n_{j}\rightarrow \infty }}}%
\frac{\overline{\boldsymbol{\mu }_{i}^{T}\mathbf{\Sigma }^{-1}\boldsymbol{%
\mu }_{i}}}{n_{j}}+\lim_{\substack{ {n_{i}\rightarrow \infty }}}\frac{%
\overline{\boldsymbol{\mu }_{j}^{T}\mathbf{\Sigma }^{-1}\boldsymbol{\mu }_{j}%
}}{n_{i}}=0\,.
\end{aligned}
\label{varG}
\end{equation}%
From the Cauchy-Schwarz inequality $(\text{Cov}[x,y]\!\leq\! \sqrt{\text{Var}%
[x]\text{Var}[y]})$, $\text{Cov}_{S_{n}}[y_{i},z_{k}|\boldsymbol{\mu }]{%
\overset{K}{\rightarrow }}0$, $\text{Cov}_{S_{n}}[y_{i},z_{ij}|\boldsymbol{%
\mu }]{\overset{K}{\rightarrow }}0$, and $\text{Cov}_{S_{n}}[z_{i},z_{ij}|%
\boldsymbol{\mu }]\;{\overset{K}{\rightarrow }}\;0$ for $i,j,k=0,1$, $i\neq
j $, Furthermore, $\frac{n_{i}-\nu _{i}}{2(n_{i}+\nu _{i})}\;{\overset{K}{%
\rightarrow }}\;\frac{1-\gamma _{i}}{2(1+\gamma _{i})}$ and $\frac{\nu _{i}}{%
n_{i}+\nu _{i}}\;{\overset{K}{\rightarrow }}\;\frac{\gamma _{i}}{1+\gamma
_{i}}$. Putting this together and following the same approach for $\hat{G}%
_{1}^{B}$ yields Var$_{S_{n}}[\hat{G}_{i}^{B}|\boldsymbol{\mu }]{\overset{K}{%
\rightarrow }}0$. In general (via Chebyshev's inequality), $\lim_{\substack{ 
{n\rightarrow \infty }}}$Var$[X_{n}]=0$ implies convergence in probability
of $X_{n}$ to $\lim_{\substack{ {n\rightarrow \infty }}}E[X_{n}]$. Hence,
since Var$_{S_{n}}[\hat{G}_{i}^{B}|\boldsymbol{\mu }]{\overset{K}{%
\rightarrow }}0$, for $i,j=0,1$ and $i\neq j$, 
\begin{equation}
\begin{aligned}
& \underset{\text{b.k.a.c.}}{\operatorname{plim}}\;\hat{G}_{i}^{B}|\boldsymbol{\mu }=%
\underset{\text{b.k.a.c.}}{\operatorname{lim}}\;E_{S_{n}}[\hat{G}_{i}^{B}|%
\boldsymbol{\mu }]=(-1)^{i}\Big[\frac{1}{2}\left( \overline{\boldsymbol{\mu }%
_{j}^{T}\mathbf{\Sigma }^{-1}\boldsymbol{\mu }_{j}}+J_{j}\right) + \\
& \frac{\gamma _{i}(\overline{\mathbf{m}_{i}^{T}\mathbf{\Sigma }^{-1}%
\boldsymbol{\mu }_{i}}-\overline{\mathbf{m}_{i}^{T}\mathbf{\Sigma }^{-1}%
\boldsymbol{\mu }_{j}})}{1+\gamma _{i}}+\frac{1-\gamma _{i}}{2(1+\gamma _{i})%
}\left( \overline{\boldsymbol{\mu }_{i}^{T}\mathbf{\Sigma }^{-1}\boldsymbol{%
\mu }_{i}}+J_{i}\right) -\overline{\boldsymbol{\mu }_{i}^{T}\mathbf{\Sigma }%
^{-1}\boldsymbol{\mu }_{j}}\left( \frac{1-\gamma _{i}}{2(1+\gamma _{i})}+%
\frac{1}{2}\right) \Big]{=}G_{i}^{B}.
\end{aligned}
\label{tyty}
\end{equation}

Now let 
\begin{equation}
\hat{D}_{i}\,=\,\frac{\nu _{i}^{\ast }+1}{\nu _{i}^{\ast }}(\bar{\mathbf{x}}%
_{0}-\bar{\mathbf{x}}_{1})^{T}\mathbf{\Sigma }^{-1}(\bar{\mathbf{x}}_{0}-%
\bar{\mathbf{x}}_{1})=\frac{\nu _{i}^{\ast }+1}{\nu _{i}^{\ast }}\hat{\delta}%
^{2},  \label{Dtrue}
\end{equation}%
where $\hat{\delta}^{2}=(\bar{\mathbf{x}}_{0}-\bar{\mathbf{x}}_{1})^{T}%
\mathbf{\Sigma }^{-1}(\bar{\mathbf{x}}_{0}-\bar{\mathbf{x}}_{1})$. Similar
to deriving (\ref{varG}) via the variance of quadratic forms of Gaussian
variables, we can show 
\begin{equation}
\text{Var}_{S_{n}}[\hat{\delta}^{2}|\boldsymbol{\mu }]=4\delta _{\boldsymbol{%
\mu }}^{2}\left( \frac{1}{n_{0}}+\frac{1}{n_{1}}\right) +2p\left( \frac{1}{%
n_{0}}+\frac{1}{n_{1}}\right) ^{2}.  \label{xhabma}
\end{equation}%
Thus, 
\begin{equation}
\begin{aligned}
\text{Var}_{S_{n}}[\hat{D}_{i}|\boldsymbol{\mu }]=\left( \frac{\nu
_{i}^{\ast }+1}{\nu _{i}^{\ast }}\right) ^{2}\text{Var}_{S_{n}}[\hat{\delta}%
^{2}|\boldsymbol{\mu }]{\overset{K}{\rightarrow }}0.
\end{aligned}
\label{cnjasa}
\end{equation}%
As before, from Chebyshev's inequality it follows that 
\begin{equation}
\underset{\text{b.k.a.c.}}{\operatorname{plim}}\;\hat{D}_{i}|\boldsymbol{\mu }=%
\underset{\text{b.k.a.c.}}{\operatorname{lim}}\;E_{S_{n}}[\hat{D}_{i}|\boldsymbol{%
\mu }]{=}D.
\end{equation}
By the Continuous Mapping Theorem (continuous functions preserve convergence
in probability), 
\begin{equation}
\begin{aligned}
& \underset{\text{b.k.a.c.}}{\operatorname{plim}}\;{\hat{\varepsilon}}_{i}^{B}|%
\boldsymbol{\mu }=\underset{\text{b.k.a.c.}}{\operatorname{plim}}\;\Phi \left(
(-1)^{i}\;\frac{-\hat{G}_{i}^{B}+c}{\sqrt{\hat{D}_{i}}}\right) |\boldsymbol{%
\mu }\!=\!\Phi \left( \underset{\text{b.k.a.c.}}{\operatorname{plim}}(-1)^{i}\;\frac{%
-\hat{G}_{i}^{B}+c}{\sqrt{\hat{D}_{i}}}|\boldsymbol{\mu }\right) \!=\!\Phi
\left( (-1)^{i}\;\frac{-G_{i}^{B}+c}{\sqrt{D}}\right) .
\end{aligned}%
\label{bskabA}
\end{equation}%
From (\ref{bskabA}) we have
\begin{equation}
\begin{aligned}
(-1)^{i}\;\frac{-\hat{G}_{i}^{B}+c}{\sqrt{\hat{D}_{i}}}|\boldsymbol{\mu }\overset{D}{\to} Z_i \sim \delta\left(z-(-1)^{i}\;\frac{-G_{i}^{B}+c}{\sqrt{D_i}}\right)
\end{aligned}%
\end{equation}%
with $\delta(.)$ being the delta function and $\overset{D}{\to}$ shows convergence in distribution. 
Boundedness and continuity of $\Phi (.)$ along with (2) allow one to apply Helly-Bray lemma \cite{SenSing:93} to write

\begin{equation}
\begin{aligned}
& \underset{\text{b.k.a.c.}}{\operatorname{lim}}E_{S_{n}}[{\hat{\varepsilon}}%
_{i}^{B}|\boldsymbol{\mu }]=\underset{\text{b.k.a.c.}}{\operatorname{lim}}\;E_{S_{n}}%
\left[ \Phi \left((-1)^{i}\; \frac{-\hat{G}_{i}^{B}+c}{\sqrt{\hat{D}_{i}}}%
\right) |\boldsymbol{\mu }\right] \\
& =E_{Z_i}\left[ \Phi \left( 
z
\right) \right] =\Phi
\left( (-1)^{i}\;\frac{-G_{i}^{B}+c}{\sqrt{D_i}}\right)=\underset{\text{b.k.a.c.}}{\operatorname{plim}}\;{\hat{\varepsilon}}%
_{i}^{B}|\boldsymbol{\mu }.\text{ \ \ \ \ \ \ }\blacksquare
\end{aligned}%
\end{equation}

\section*{Proof of Theorem 2}

We first prove that $\text{Var}_{\boldsymbol{\mu },S_{n}}(\hat{G}_{0}^{B}){%
\overset{K}{\rightarrow }}0$ with $\hat{G}_{0}^{B}$ defined in (\ref{ghat0}%
). To do so we use 
\begin{equation}
\begin{aligned}
\text{Var}_{\boldsymbol{\mu },S_{n}}[\hat{G}_{0}^{B}]=\text{Var}_{%
\boldsymbol{\mu }}\left[ E_{S_{n}}[\hat{G}_{0}^{B}|\boldsymbol{\mu }]\right]
+E_{\boldsymbol{\mu }}\left[ \text{Var}_{S_{n}}[\hat{G}_{0}^{B}|\boldsymbol{%
\mu }]\right] .
\end{aligned}
\label{condv}
\end{equation}%
To compute the first term on the right hand side, we have 
\begin{equation}
\begin{aligned}
E_{S_{n}}[\hat{G}_{0}^{B}|\boldsymbol{\mu }]& =\,\frac{\nu _{0}\mathbf{m}%
_{0}^{T}}{n_{0}+\nu _{0}}\mathbf{\Sigma }^{-1}(\boldsymbol{\mu }_{0}-%
\boldsymbol{\mu }_{1})+\frac{n_{0}-\nu _{0}}{2(n_{0}+\nu _{0})}\left( 
\boldsymbol{\mu }_{0}^{T}\mathbf{\Sigma }^{-1}\boldsymbol{\mu }_{0}+\frac{p}{%
n_{0}}\right) \\
& +\frac{1}{2}\left( \boldsymbol{\mu }_{1}^{T}\mathbf{\Sigma }^{-1}%
\boldsymbol{\mu }_{1}+\frac{p}{n_{1}}\right) -\boldsymbol{\mu }_{0}^{T}%
\mathbf{\Sigma }^{-1}\boldsymbol{\mu }_{1}\left( \frac{n_{0}}{n_{0}+\nu _{0}}%
\right) .
\end{aligned}%
\end{equation}%
For $i,j=0,1$ and $i\neq j$ define the following random variables: 
\begin{equation}
\begin{aligned}
y_{i}^{\prime }=\mathbf{m}_{i}^{T}\mathbf{\Sigma }^{-1}(\boldsymbol{\mu }%
_{0}-\boldsymbol{\mu }_{1}),\mathrm{\ \ \ }z_{i}^{\prime }=\boldsymbol{\mu }%
_{i}^{T}\mathbf{\Sigma }^{-1}\boldsymbol{\mu }_{i},\mathrm{\ \ \ }%
z_{ij}^{\prime }=\boldsymbol{\mu }_{i}^{T}\mathbf{\Sigma }^{-1}\boldsymbol{%
\mu }_{j}.
\end{aligned}
\label{fff}
\end{equation}%
The variables defined in (\ref{fff}) can be obtained by replacing $\bar{%
\mathbf{x}}_{i}$'s with $\boldsymbol{\mu }_{i}$'s in (\ref{ggg}) and $\bar{%
\mathbf{x}}_{i}\sim N(\boldsymbol{\mu }_{i},\mathbf{\Sigma }/n_{i})$ and $%
\boldsymbol{\mu }_{i}\sim N(\mathbf{m}_{i},\mathbf{\Sigma }/\nu _{i})$.
Replacing $\boldsymbol{\mu }_{i}$ with $\mathbf{m}_{i}$ and $n_{i}$ with $%
\nu _{i}$ in (\ref{varGj}) and (\ref{varG}) yields 
\begin{equation}
\begin{aligned}
& \text{Var}_{\boldsymbol{\mu }}(y_{i}^{\prime })=\mathbf{m}_{i}^{T}(\frac{%
\mathbf{\Sigma }^{-1}}{\nu _{0}}+\frac{\mathbf{\Sigma }^{-1}}{\nu _{1}})%
\mathbf{m}_{i}\;{\overset{K}{\rightarrow }}\;0, \quad \text{Var}_{\boldsymbol{\mu }}(z_{i}^{\prime })=2\frac{p}{\nu _{i}^{2}}+4%
\frac{\mathbf{m}_{i}^{T}\mathbf{\Sigma }^{-1}\mathbf{m}_{i}}{\nu _{i}}\;{%
\overset{K}{\rightarrow }}\;0, \\
& \text{Var}_{\boldsymbol{\mu }}(z_{ij}^{\prime })=\frac{p}{\nu _{i}\nu _{j}}%
+\frac{\mathbf{m}_{i}^{T}\mathbf{\Sigma }^{-1}\mathbf{m}_{i}}{\nu _{j}}+%
\frac{\mathbf{m}_{j}^{T}\mathbf{\Sigma }^{-1}\mathbf{m}_{j}}{\nu _{i}}\;{%
\overset{K}{\rightarrow }}\;0.
\end{aligned}
\label{varGp}
\end{equation}%
By Cauchy-Schwarz, $\text{Cov}_{\boldsymbol{\mu }}(y_{i}^{\prime
},z_{k}^{\prime })\;{\overset{K}{\rightarrow }}\;0$, $\text{Cov}_{%
\boldsymbol{\mu }}(y_{i}^{\prime },z_{ij}^{\prime })\;{\overset{K}{%
\rightarrow }}\;0$, and $\text{Cov}_{\boldsymbol{\mu }}(z_{i}^{\prime
},z_{ij}^{\prime })\;{\overset{K}{\rightarrow }}\;0$. Hence, \newline
$\text{Var}_{\boldsymbol{\mu }}\left[ E_{S_{n}}[\hat{G}_{0}^{B}|\boldsymbol{%
\mu }]\right] {\overset{K}{\rightarrow }}\;0$.

Now consider the second term on the right hand side of (\ref{condv}). The
covariance of a function of Gaussian random variables can be computed from
results of \cite{Ullah1}. For instance, 
\begin{equation}
\text{Cov}_{S_{n}}[\mathbf{a}^{T}\bar{\mathbf{x}}_{i},\;\bar{\mathbf{x}}%
_{i}^{T}\mathbf{\Sigma }^{-1}\bar{\mathbf{x}}_{i}|\boldsymbol{\mu }]=\frac{2%
}{n_{i}}\mathbf{a}^{T}\boldsymbol{\mu }_{i}\,.  \label{kasa}
\end{equation}%
From (\ref{kasa}) and the independence of $\bar{\mathbf{x}}_{0}$ and $\bar{%
\mathbf{x}}_{1}$, 
\begin{equation}
\text{Cov}_{S_{n}}[\bar{\mathbf{x}}_{j}^{T}\mathbf{\Sigma }^{-1}\bar{\mathbf{%
x}}_{i},\;\bar{\mathbf{x}}_{i}^{T}\mathbf{\Sigma }^{-1}\bar{\mathbf{x}}_{i}|%
\boldsymbol{\mu }]=\frac{2}{n_{i}}\boldsymbol{\mu }_{j}^{T}\mathbf{\Sigma }%
^{-1}\boldsymbol{\mu }_{i},\;\;i\neq j  \label{kasa1}
\end{equation}%
Via (\ref{varGp}), (\ref{kasa}), and (\ref{kasa1}), the inner variance in
the second term on the right hand side of (\ref{condv}) is 
\begin{equation}
\begin{aligned}
& \text{Var}_{S_{n}}[\hat{G}_{0}^{B}|\boldsymbol{\mu }]=\frac{(n_{0}-\nu
_{0})^{2}p}{2n_{0}^{2}(n_{0}+\nu _{0})^{2}}+\frac{n_{0}p}{n_{1}(n_{0}+\nu
_{0})^{2}}+\frac{p}{2n_{1}^{2}} \\
& +\left( \frac{(n_{0}-\nu _{0})^{2}}{n_{0}(n_{0}+\nu _{0})^{2}}+\frac{%
n_{0}^{2}}{n_{1}(n_{0}+\nu _{0})^{2}}\right) \boldsymbol{\mu }_{0}^{T}%
\mathbf{\Sigma }^{-1}\boldsymbol{\mu }_{0}-2\left( \frac{n_{0}-\nu _{0}}{%
(n_{0}+\nu _{0})^{2}}+\frac{n_{0}}{n_{1}(n_{0}+\nu _{0})}\right) \boldsymbol{%
\mu }_{0}^{T}\mathbf{\Sigma }^{-1}\boldsymbol{\mu }_{1} \\
& +2\left( \frac{\nu _{0}(n_{0}-\nu _{0})}{n_{0}(n_{0}+\nu _{0})^{2}}+\frac{%
\nu _{0}n_{0}}{n_{1}(n_{0}+\nu _{0})^{2}}\right) \mathbf{m}_{0}^{T}\mathbf{%
\Sigma }^{-1}\boldsymbol{\mu }_{0}-2\left( \frac{\nu _{0}}{(n_{0}+\nu
_{0})^{2}}+\frac{\nu _{0}}{n_{1}(n_{0}+\nu _{0})}\right) \mathbf{m}_{0}^{T}%
\mathbf{\Sigma }^{-1}\boldsymbol{\mu }_{1}+ \\
& \left( \frac{n_{0}}{(n_{0}+\nu _{0})^{2}}+\frac{1}{n_{1}}\right) 
\boldsymbol{\mu }_{1}^{T}\mathbf{\Sigma }^{-1}\boldsymbol{\mu }_{1}+\frac{%
\nu _{0}^{2}}{(n_{0}+\nu _{0})^{2}}\left( \frac{1}{n_{0}}+\frac{1}{n_{1}}%
\right) \mathbf{m}_{0}^{T}\mathbf{\Sigma }^{-1}\mathbf{m}_{0}.
\end{aligned}
\label{hdhs}
\end{equation}%
Now, again from the results of \cite{Ullah1}, 
\begin{equation}
\begin{aligned}
& E_{\boldsymbol{\mu }}[\boldsymbol{\mu }_{i}^{T}\mathbf{\Sigma }^{-1}%
\boldsymbol{\mu }_{i}]=\mathbf{m}_{i}^{T}\mathbf{\Sigma }^{-1}\mathbf{m}_{i}+%
\frac{p}{\nu _{i}},\\&  E_{\boldsymbol{\mu }}[\boldsymbol{\mu }_{i}^{T}\mathbf{\Sigma }^{-1}%
\boldsymbol{\mu }_{j}]=\mathbf{m}_{i}^{T}\mathbf{\Sigma }^{-1}\mathbf{m}%
_{j},i\neq j.
\end{aligned}
\label{bkashbkaU}
\end{equation}%
From (\ref{hdhs}) and (\ref{bkashbkaU}), some algebraic manipulations yield 
\begin{equation}
\begin{aligned}
& E_{\boldsymbol{\mu }}\left[ \text{Var}_{S_{n}}[\hat{G}_{0}^{B}|\boldsymbol{%
\mu }]\right] =\frac{(n_{0}-\nu _{0})^{2}p}{2n_{0}^{2}(n_{0}+\nu _{0})^{2}}+%
\frac{n_{0}p}{n_{1}(n_{0}+\nu _{0})^{2}} \\
& +\frac{p}{2n_{1}^{2}}+\left( \frac{(n_{0}-\nu _{0})^{2}}{n_{0}(n_{0}+\nu
_{0})^{2}}+\frac{n_{0}^{2}}{n_{1}(n_{0}+\nu _{0})^{2}}\right) \frac{p}{\nu
_{0}} +\left( \frac{n_{0}}{(n_{0}+\nu _{0})^{2}}+\frac{1}{n_{1}}\right) 
\frac{p}{\nu _{1}}+\left( \frac{n_{0}}{(n_{0}+\nu _{0})^{2}}+\frac{1}{n_{1}}%
\right) \Delta _{\mathbf{m}}^{2}.
\end{aligned}
\label{jeyxn}
\end{equation}%
From (\ref{jeyxn}) we see that $E_{\boldsymbol{\mu }}\big[\text{Var}_{S_{n}}[%
\hat{G}_{0}^{B}|\boldsymbol{\mu }]\big]{\overset{K}{\rightarrow }}\;0$. In
sum, $\text{Var}_{\boldsymbol{\mu },{S_{n}}}[\hat{G}_{0}^{B}]{\overset{K}{%
\rightarrow }}\;0$ and similar to the use Chebyshev's inequality in the
proof of Theorem \ref{thm-m1}, we get 
\begin{equation}
\underset{\text{b.k.a.c.}}{\operatorname{plim}}\;\hat{G}_{i}^{B}=\underset{\text{%
b.k.a.c.}}{\operatorname{lim}}\;E_{\boldsymbol{\mu },{S_{n}}}[\hat{G}_{i}^{B}]{%
\overset{\Delta }{\;=\;}}H_{i},  \label{tytyr}
\end{equation}%
with $H_{i}$ defined in (\ref{njzzp}).

On the other hand, for $\hat{D}_{i}$ defined in (\ref{Dtrue}) we can write 
\begin{equation}
\begin{aligned}
\text{Var}_{\boldsymbol{\mu },S_{n}}[\hat{D}_{i}]=\text{Var}_{\boldsymbol{%
\mu }}\left[ E_{S_{n}}[\hat{D}_{i}|\boldsymbol{\mu }]\right] +E_{\boldsymbol{%
\mu }}\left[ \text{Var}_{S_{n}}[\hat{D}_{i}|\boldsymbol{\mu }]\right] .
\end{aligned}
\label{condvD}
\end{equation}%
From similar expressions as in (\ref{bkashbkaU}) for $\bar{\mathbf{x}}%
_{i}^{T}\mathbf{\Sigma }^{-1}\bar{\mathbf{x}}_{j}$, we get $E_{S_{n}}[\hat{%
\delta}^{2}]=\delta _{\boldsymbol{\mu }}^{2}+\frac{p}{n_{0}}+\frac{p}{n_{1}}$%
. Moreover, $\text{Var}_{\boldsymbol{\mu }}[\delta _{\boldsymbol{\mu }}^{2}]$
is obtained from (\ref{xhabma}) by replacing $n_{i}$ with $\nu _{i}$, and $%
\delta _{\boldsymbol{\mu }}^{2}$ with $\Delta _{\mathbf{m}}^{2}$. Thus, from
(\ref{Dtrue}), 
\begin{equation}
\begin{aligned}
\text{Var}_{\boldsymbol{\mu }}\left[ E_{S_{n}}[\hat{D}_{i}|\boldsymbol{\mu }]%
\right] =\left( \frac{\nu _{i}^{\ast }+1}{\nu _{i}^{\ast }}\right) ^{2}\bigg[%
4\Delta _{\mathbf{m}}^{2}\left( \frac{1}{\nu _{0}}+\frac{1}{\nu _{1}}\right)
+2p\left( \frac{1}{\nu _{0}}+\frac{1}{\nu _{1}}\right) ^{2}\bigg]{\overset{K}%
{\rightarrow }}0.
\end{aligned}
\label{cnjasak}
\end{equation}%
Furthermore, since $E_{\boldsymbol{\mu }}[\delta _{\boldsymbol{\mu }%
}^{2}]=\Delta _{\mathbf{m}}^{2}+\frac{p}{\nu _{0}}+\frac{p}{\nu _{1}}$, from
(\ref{cnjasa}), 
\begin{equation}
\begin{aligned}
E_{\boldsymbol{\mu }}\left[ \text{Var}_{S_{n}}[\hat{D}_{i}|\boldsymbol{\mu }]%
\right] & =\left( \frac{\nu _{i}^{\ast }+1}{\nu _{i}^{\ast }}\right) ^{2} %
\left[ 4(\Delta _{\mathbf{m}}^{2}+\frac{p}{\nu _{0}}+\frac{p}{\nu _{1}}%
)\left( \frac{1}{n_{0}}+\frac{1}{n_{1}}\right) +2p\left( \frac{1}{n_{0}}+%
\frac{1}{n_{1}}\right) ^{2}\right] {\overset{K}{\rightarrow }}0.
\end{aligned}
\label{mshdh}
\end{equation}%
Hence, $\text{Var}_{\boldsymbol{\mu },S_{n}}[\hat{D}_{i}]{\overset{K}{%
\rightarrow }}0$ and, similar to (\ref{tytyr}), we obtain 
\begin{equation}
\underset{\text{b.k.a.c.}}{\operatorname{plim}}\;\hat{D}_{i}=\underset{\text{b.k.a.c.%
}}{\operatorname{lim}}\;E_{\boldsymbol{\mu },S_{n}}[\hat{D}_{0}]=\underset{\text{%
b.k.a.c.}}{\operatorname{lim}}\;E_{\boldsymbol{\mu },S_{n}}[\hat{D}_{1}]{\overset{%
\Delta }{\;=\;}}F,  \label{msanxs}
\end{equation}%
with $F$ defined in (\ref{njzzp}). Similar to the proof of Theorem \ref%
{thm-m1}, by using the Continuous Mapping Theorem and the Helly-Bray lemma, we can show that 
\begin{equation}
\begin{aligned}
& \lim_{\substack{ \text{b.k.a.c.}  \\ }}E_{\boldsymbol{\mu },S_{n}}\left[ 
\hat{\varepsilon}_{i}^{B}\right] =\underset{\text{b.k.a.c.}}{\operatorname{lim}}E_{%
\boldsymbol{\mu },S_{n}}\left[ \Phi \left((-1)^{i}\; \frac{{-\hat{G}}%
_{i}^{B}+c }{\sqrt{{\hat{D}}}}\right) \right] \\
& =\Phi \left( (-1)^{i}\;\frac{-H_{i}+c}{\sqrt{F}}\right) ,
\end{aligned}
\label{mdns}
\end{equation}%
and the result follows. \ \ $\blacksquare$

\section*{Proof of Theorem 3}

We start from 
\begin{equation}
\begin{aligned}
E_{S_{n}}[(\hat{\varepsilon}_{0}^{B})^{2}|\boldsymbol{\mu }]=E_{S_{n}}\bigg[P%
\Big(U_{0}(\bar{\mathbf{x}}_{0},\bar{\mathbf{x}}_{1},\mathbf{z})\leq
c\,,U_{0}(\bar{\mathbf{x}}_{0},\bar{\mathbf{x}}_{1},\mathbf{z}^{\prime
})\leq c\,|\bar{\mathbf{x}}_{0},\bar{\mathbf{x}}_{1},\mathbf{z}\in \Psi _{0},%
\mathbf{z}^{\prime }\in \Psi _{0},\boldsymbol{\mu }\Big)\bigg],
\end{aligned}%
\end{equation}%
which was shown in (\ref{kjxbkasU}). Here we characterize the conditional
probability inside $E_{S_{n}}[.]$. From the independence of $\mathbf{z}$, $%
\mathbf{z}^{\prime }$, $\bar{\mathbf{x}}_{0}$, and $\bar{\mathbf{x}}_{1}$, 
\begin{equation}
\begin{aligned}
\begin{bmatrix}
U_{0}(\bar{\mathbf{x}}_{0},\bar{\mathbf{x}}_{1},\mathbf{z}) \\ 
U_{0}(\bar{\mathbf{x}}_{0},\bar{\mathbf{x}}_{1},\mathbf{z}^{\prime })%
\end{bmatrix}%
\mid \;\bar{\mathbf{x}}_{0},\bar{\mathbf{x}}_{1},\mathbf{z}\in \Psi _{0},%
\mathbf{z}^{\prime }\in \Psi _{0},\boldsymbol{\mu }\sim N\left( 
\begin{bmatrix}
\hat{G}_{0}^{B} \\ 
\hat{G}_{0}^{B}%
\end{bmatrix}%
\,,%
\begin{bmatrix}
\hat{D} & 0 \\ 
0 & \hat{D}%
\end{bmatrix}%
\,\right) ,
\end{aligned}
\label{lkjhgfdsar}
\end{equation}%
where here $N\left( .\,,.\,\right) $ denotes the bivariate Gaussian density
function and $\hat{G}_{0}^{B}$ and $\hat{D}$ are defined in (\ref{ghat0})
and (\ref{Dtrue}). Thus, 
\begin{equation}
E_{S_{n}}[(\hat{\varepsilon}_{0}^{B})^{2}|\boldsymbol{\mu }]=\left[ \Phi
\left( \frac{-\hat{G}_{0}^{B}+c}{\sqrt{\hat{D}}}\right) \right] ^{2}|%
\boldsymbol{\mu }.
\end{equation}%
Similar to the proof of Theorem \ref{thm-m1}, we get 
\begin{equation}
\begin{aligned}
\lim_{\substack{ \text{b.k.a.c.}  \\ }}E_{S_{n}}[(\hat{\varepsilon}%
_{i}^{B})^{2}|\boldsymbol{\mu }]=\underset{\text{b.k.a.c.}}{\operatorname{plim}}\;(%
\hat{\varepsilon}_{i}^{B})^{2}|\boldsymbol{\mu }=\left( \lim_{\substack{ 
\text{b.k.a.c.}  \\ }}E_{S_{n}}[\hat{\varepsilon}_{i}^{B}|\boldsymbol{\mu }%
]\right) ^{2}=\left[ \Phi \left( (-1)^{i}\;\frac{-G_{i}^{B}+c}{\sqrt{D}}%
\right) \right] ^{2}.
\end{aligned}%
\end{equation}%
Similarly, we obtain $\lim_{b.k.a.c.}E[\hat{\varepsilon}_{0}^{B}\hat{%
\varepsilon}_{1}^{B}]=$ $\lim_{b.k.a.c.}\hat{\varepsilon}_{0}^{B}{%
\varepsilon }_{1}^{B}$, and the results follow. \ \ $\blacksquare$


\end{document}